\documentclass[journal]{IEEEtran}
\usepackage{cite}
\usepackage{pifont}
\newcommand{\cmark}{\ding{51}}%
\ifCLASSINFOpdf
  \usepackage[pdftex]{graphicx}
	\usepackage{epstopdf}
\else
  \usepackage[dvips]{graphicx}
\fi
\usepackage[cmex10]{amsmath}
\usepackage{algorithmic}
\usepackage{amssymb}
\usepackage{array}
\usepackage{mdwmath}
\usepackage{mdwtab}
\usepackage{eqparbox}
\usepackage{multirow}
\usepackage{fixltx2e}
\usepackage{enumerate}
\usepackage{url}
\usepackage{tabularx}
\begin{document}
%
\title{VIDOSAT: High-dimensional Sparsifying Transform Learning for Online Video Denoising}
\author{Bihan~Wen, ~\IEEEmembership{Student Member,~IEEE,}~Saiprasad~Ravishankar, ~\IEEEmembership{Member,~IEEE,} and~Yoram~Bresler,~\IEEEmembership{Fellow,~IEEE}
\thanks{This work was supported in part by the National Science Foundation (NSF) under grant CCF-1320953. Saiprasad Ravishankar was supported in part by the following grants: ONR grant N00014-15-1-2141, DARPA Young Faculty Award D14AP00086, ARO MURI grants W911NF-11-1-0391 and 2015-05174-05, and a UM-SJTU seed grant.}
\thanks{B. Wen, and Y. Bresler are with the Department of Electrical and Computer Engineering and the Coordinated Science Laboratory, University of Illinois, Urbana-Champaign, IL, 61801 USA e-mail: (bwen3, ybresler)@illinois.edu.}
\thanks{S. Ravishankar is with the Department of Electrical Engineering and Computer Science, University of Michigan, Ann Arbor, MI 48109, USA e-mail: ravisha@umich.edu.}}

\maketitle

\begin{abstract}
Techniques exploiting the sparsity of images in a transform domain have been effective for various applications in image and video processing. 
Transform learning methods involve cheap computations and have been demonstrated to perform well in applications such as image denoising and medical image reconstruction. 
Recently, we proposed methods for online learning of sparsifying transforms from streaming signals, which enjoy good convergence guarantees, and involve lower computational costs than online synthesis dictionary learning. 
In this work, we apply online transform learning to video denoising. 
We present a novel framework for online video denoising based on high-dimensional sparsifying transform learning for spatio-temporal patches. 
The patches are constructed either from corresponding 2D patches in successive frames or using an online block matching technique.
The proposed online video denoising requires little memory, and offers efficient processing.
Numerical experiments compare the performance to the proposed video denoising scheme but fixing the transform to be 3D DCT, as well as prior schemes such as dictionary learning-based schemes, and the state-of-the-art VBM3D and VBM4D on several video data sets, demonstrating the promising performance of the proposed methods.
\end{abstract}


\IEEEpeerreviewmaketitle

\section{Introduction} \label{intro}
Recent techniques in image and video processing make use of sophisticated models of signals and images.
Various properties such as sparsity, low-rank, etc., have been exploited in inverse problems such as video denoising, or other dynamic image reconstruction problems such as magnetic resonance imaging or positron emission tomography \cite{yoon2014motion,kim2015dynamic}.
Adaptive or data-driven models and approaches are gaining increasing interest. This work presents novel online data-driven video denoising techniques based on learning sparsifying transforms for appropriately constructed spatio-temporal patches of videos. This new framework provides high quality video restoration from highly corrupted data.
In the following, we briefly review the background on video denoising and sparsifying transform learning, before discussing the contributions of this work.

\subsection{Video Denoising}
Denoising is one of the most important problems in video processing. The ubiquitous use of relatively low-quality smart phone cameras has also led to the increasing importance of video denoising. 
Recovering high-quality video from noisy footage also improves robustness in high-level vision tasks \cite{Mairal2009,bengio2010stacked}.

Though image denoising algorithms, such as the popular BM3D method \cite{Dabov2007} can be applied to each video frame independently, most of the video denoising techniques (or more generally, methods for reconstructing dynamic data from measurements \cite{ota1,sravbrijefraj}) exploit the spatio-temporal correlation in dynamic image sequences.
Natural videos have local structures that are sparse or compressible in some transform domain, or in certain dictionaries, e.g., discrete cosince transform (DCT) \cite{rusanovskyy} and wavelets \cite{rajpoot}. 
Prior works exploited this fact and proposed video (or high-dimensional data) denoising algorithms based on 
adaptive sparse approximation \cite{elad2010sksvd} or Wiener filtering \cite{vbm3d}.
Videos also typically involve various kinds of motion or dynamics in the scene, e.g., moving objects or humans, rotations, etc.
Recent state-of-the-art video and image denoising algorithms utilize block matching (BM) to group local patches over space and time (to account for motion), and apply denoising jointly for such matched data \cite{Dabov2007,vbm3d,Maggioni2012}. Table \ref{Tab:compareDenoising} summarizes the key attributes of the popular and related video denoised methods, as well as the proposed methods.

\begin{figure}[!t]
\centering
\includegraphics[width=3.3in]{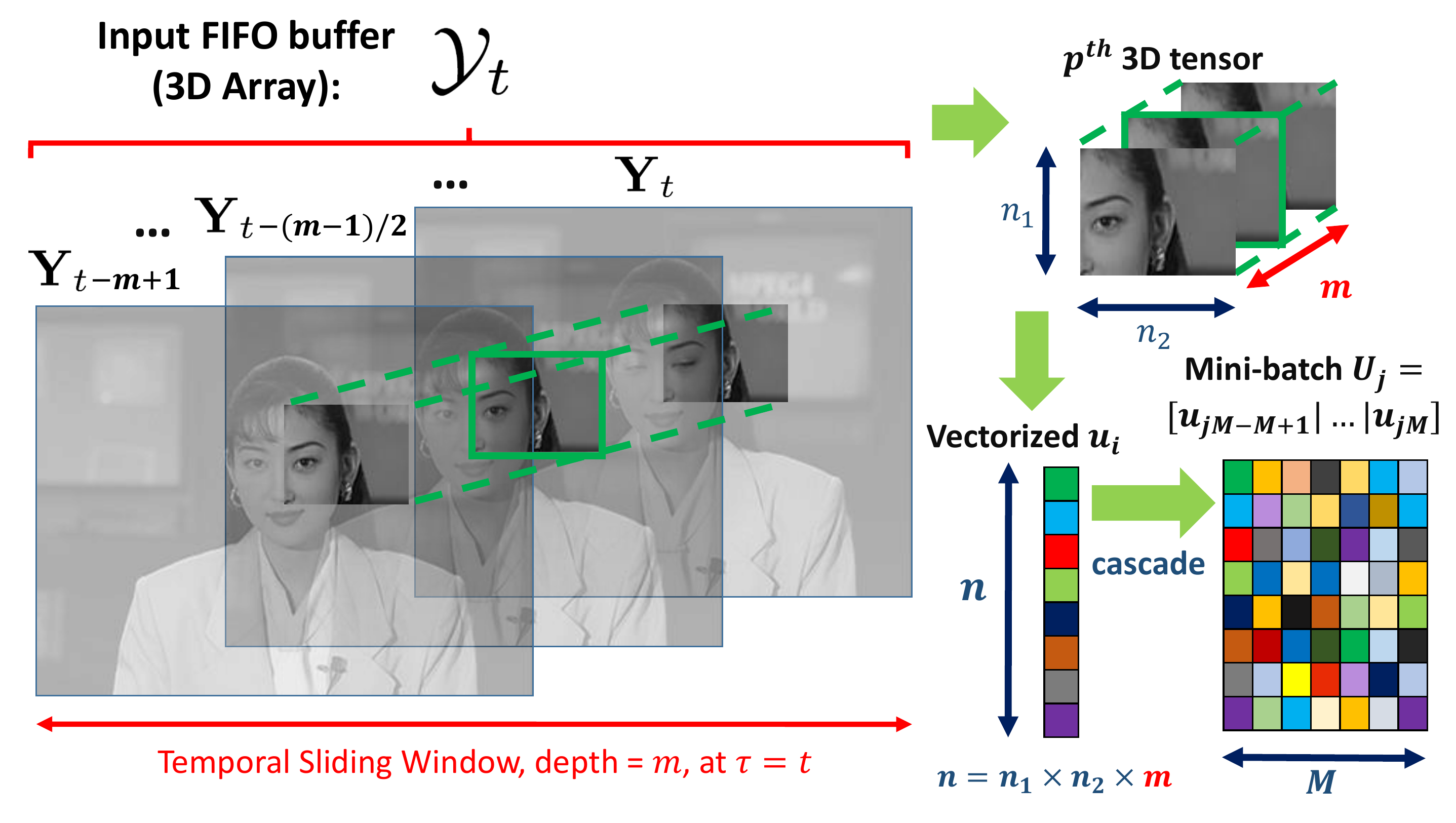}
\vspace{-0.1in}
\caption{Video streaming, tensor construction and vectorization.}
\label{extraction}
\vspace{-0.1in}
\end{figure}

\begin{table}[t!]
\centering
\fontsize{9}{12pt}\selectfont
\begin{tabular}{|c|c|c|c|c|c|}
\hline
\multirow{2}{*}{\textbf{ Methods }} & \multicolumn{3}{c|}{Sparse Signal Model} & \multirow{2}{*}{BM} & Temporal \\

\cline{2-4}
 & Fixed & Adaptive & Online &  & Correlation \\
\hline
fBM3D & \cmark & & & \cmark & \\
\hline
3D DCT & \cmark & & & & \cmark\\
\hline
sKSVD &  & \cmark &  &  & \cmark\\

\hline
VBM3D & \cmark &  &  & \cmark & \cmark \\
\hline
VBM4D & \cmark &  &  & \cmark & \cmark \\

\hline
\textbf{VIDOSAT} &  & \cmark & \cmark & & \cmark\\
\hline
\textbf{VIDOSAT} &  & \multirow{2}{*}{\cmark} & \multirow{2}{*}{\cmark} & \multirow{2}{*}{\cmark} & \multirow{2}{*}{\cmark}\\
-\textbf{BM} & &  & &  &\\
\hline
\end{tabular}
\caption{Comparison between video denoising methods, including fBM3D \cite{Dabov2007}, 3D DCT, sKSVD \cite{elad2010sksvd}, VBM3D \cite{vbm3d}, VBM4D \cite{Maggioni2012}, as well as VIDOSAT and VIDOSAT-BM prpoposed here. fBM3D is applying BM3D algorithm for denoising each frame, and the 3D DCT method is applying the VIDOSAT framework but using the fixed 3D DCT transform.}
\label{Tab:compareDenoising}
\vspace{-0.2in}
\end{table}


\vspace{-0.1in}
\subsection{Sparsifying Transform Learning}
Many of the aforementioned video denoising methods exploit sparsity in a fixed transform domain (e.g., DCT) as part of their framework. 
Several recent works have shown that the data-driven adaptation of sparse signal models (e.g., based on training signals, or directly from corrupted measurements) usually leads to high quality results (e.g., compared to fixed or analytical models) in many applications  \cite{Elad2006,elad3,protter2009image,irami,bresai,akd,sai2013tl,doubsp2l,wen2015octobos,Cai201489,zhan33,zhengsai}. 
Synthesis dictionary learning is the best-known adaptive sparse representation technique \cite{Aharon2006,Elad2006}.
However, obtaining optimal sparse representations of signals in synthesis dictionary models, known as synthesis sparse coding, is NP-hard (Non-deterministic Polynomial-time hard) in general. 
The commonly used approximate sparse coding algorithms  \cite{pati,befro,Needell2,wei} typically still involve relatively expensive computations for large-scale problems.

As an alternative, the sparsifying transform model suggests that the signal $\mathbf{u}$ is approximately sparsifiable using a transform $\mathbf{W} \in \mathbb{R}^{m \times n}$, i.e., $ \mathbf{W u = x + e} $, with $\mathbf{x} \in \mathbb{R}^{m} $ a sparse vector called the sparse code and $ \mathbf{e} $ a modeling error term in the transform domain.  
A key advantage of this model over the synthesis dictionary model, is that for a given transform $\mathbf{W}$, 
the optimal sparse code $\mathbf{x}$ of sparsity level $s$ minimizing the modeling error $\| \mathbf{e}\|_2$ is obtained \emph{exactly and cheaply} by simple thresholding of $\mathbf{W} \mathbf{u}$ to its $s$ largest magnitude components. Another advantage is that with $\mathbf{u}$ being given data, the transform model does not involve a product between $\mathbf{W}$ and unknown data, so learning  algorithms for $\mathbf{W}$ can be simpler and more reliable.
Recent works proposed learning sparsifying transforms \cite{sai2013tl,sai2015closed} with cheap algorithms that alternate between updating the sparse approximations of training signals in a transform domain using simple thresholding-based transform sparse coding, and efficiently updating the sparsifying transform.
Various properties have been found to be useful for learned transforms such as double sparsity \cite{doubsp2l}, union-of-transforms \cite{wen2015octobos},  rotation and flip invariance \cite{wen2016frist}, etc.
Transform learning-based techniques have been shown to be useful in various applications such as sparse data representations, image denoising, inpainting, segmentation, magnetic resonance imaging (MRI), and computed tomography (CT) \cite{wen2015octobos,ravishankar2015efficient,wen2016frist,Pfister2014,pfister2014tomographic,Pfister2015,dev2016ground,zhengsai,sravTCI1,zhengsaiuol}. 

In prior works on batch transform learning \cite{sai2013tl,sai2015closed,wen2015octobos,wen2016frist}, the transform was adapted using all the training data, which is efficient and comes with a convergence guarantee. 
When processing large-scale streaming data, it is also important to compute results online, or sequentially over time. 
Our recent work \cite{sai2015onlineTL,sai2015onlineTL2} proposed online transform learning, which sequentially adapts the sparsifying transform and transform-sparse coefficients for sequentially processed signals. 
This approach involves cheap computation and limited memory requirement.
Compared to popular techniques for online synthesis dictionary learning \cite{mairal2010online}, the online adaptation of sparsifying transforms allows for cheaper or exact updates \cite{sai2015onlineTL}, and is thus well suited for high-dimensional data applications.  

\begin{figure}[!t]
\centering
\includegraphics[width=3.3in]{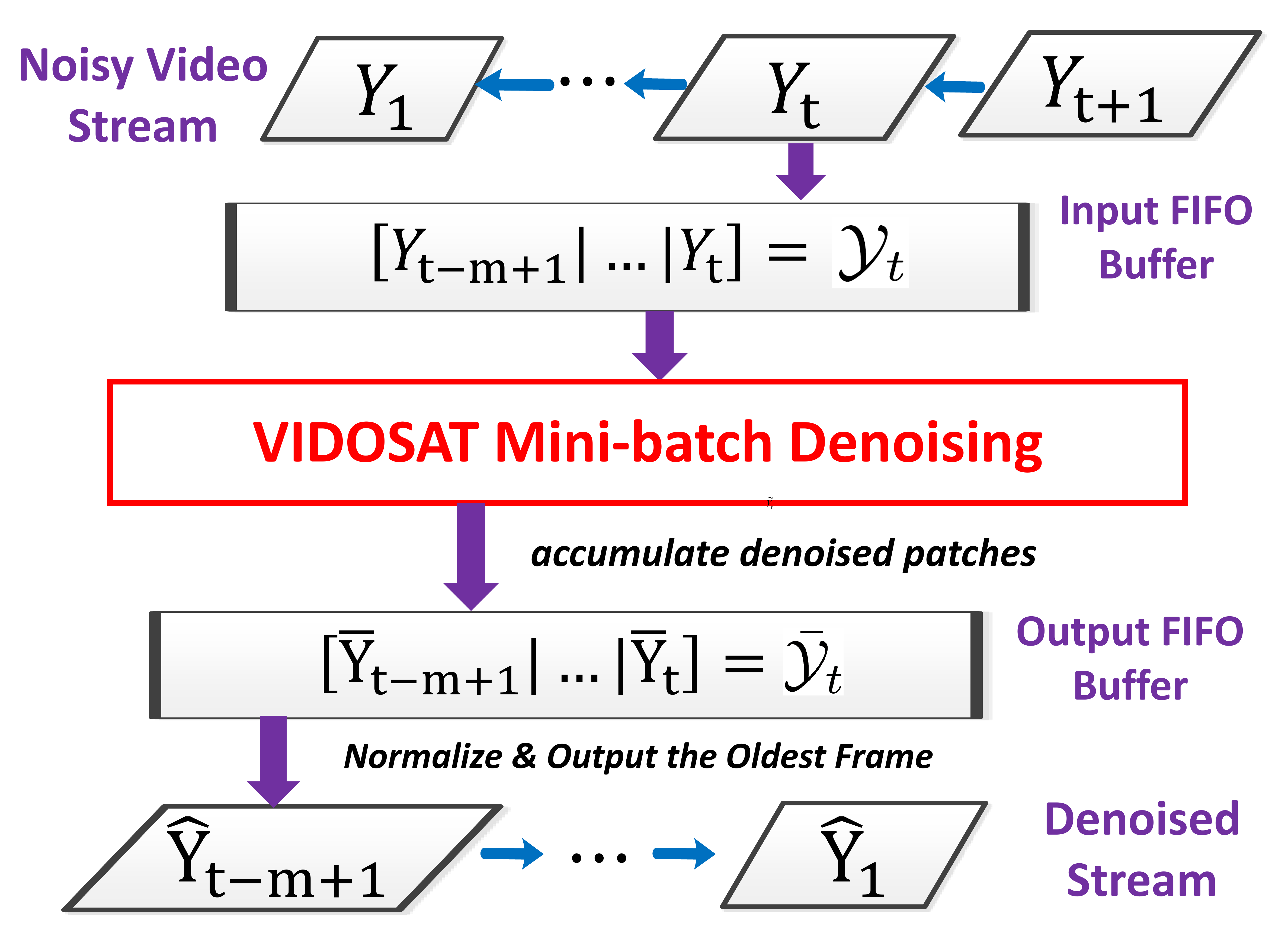}
\caption{Illustration of online video streaming and denoising framework.}
\label{frame}
\vspace{-0.05in}
\end{figure}

\vspace{-0.1in}
\subsection{Methodologies and Contributions}
While the data-driven adaptation of synthesis dictionaries for the purpose of denoising image sequences or volumetric data \cite{protter2009image,elad2010sksvd} has been studied in some recent papers, the usefulness of learned sparsifying transforms has not been explored in these applications. 
Video data typically contain correlation along the temporal dimension, which will not be captured by learning sparsifying transforms for the 2D patches of the video frames.
We focus on video denoising using high-dimensional online transform learning. We refer to our proposed framework as VIdeo Denoising by Online SpArsifying Transform learning (VIDOSAT). 
Spatio-temporal (3D) patches are constructed using local 2D patches of the corrupted video, and the sparsifying transform is adapted to these 3D patches on-the-fly.
Fig.\ref{extraction} illustrates one way of constructing the (vectorized) spatio-temporal patches or tensors from the streaming video, and Fig.\ref{frame} is a flow-chart of the proposed VIDOSAT framework.
Though we consider 3D spatio-temporal tensors formed by 2D patches for gray-scale video denoising in this work, the proposed denoising methods readily apply to higer-dimensional data (e.g., color video \cite{wen2017salt}, hyperspectral images, dynamic 3D MRI, etc) as well.

As far as we know, this is the first online video denoising method using adaptive sparse signal modeling, and the first application of high-dimensional sparsifying transform learning to spatio-temporal data. 
Our methodology and results are summarized as follows:

\begin{itemize}

 	\item The proposed video denoising framework processes noisy frames in an online, sequential fashion to produce streaming denoised video frames. 
	The algorithms require limited storage of a few video frames, and modest computation, scaling linearly with the number of pixels per frame.  As such, our methods would be able to handle high definition / high rate video enabling real-time output with controlled delay, using modest computational resources.

	
	\item The online transform learning technique exploits the spatio-temporal structure of the video tensors (patches) using adaptive 3D transform-domain sparsity to process them sequentially. 
	The denoised tensors are aggregated to reconstruct the streaming video frames.
	
	\item We evaluate the video denoising performance of the proposed algorithms for several datasets, and demonstrate their promising performance compared to several prior or related methods.
	
\end{itemize}

This paper is an extension of our previous conference work \cite{wen2015vidosat} that briefly investigated a specific VIDOSAT method. 
Compared with this earlier work, here we investigate different VIDOSAT methodologies such as involving block matching (referred to as VIDOSAT-BM).
Moreover, we provide detailed experimental results illustrating the properties of the proposed methods and their performance for several datasets, with extensive evaluation and comparison to prior or related methods.
We also demonstrate the advantages of VIDOSAT-BM over the VIDOSAT approach of \cite{wen2015vidosat}.

\subsection{Major Notations}
We use the following notations in this work.
Vectors (resp. matrices) are denoted by boldface lowercase (resp. uppercase) letters such as $\mathbf{u}$ (resp. $\mathbf{U}$). 
We use calligraphic uppercase letters (e.g., $\mathcal{U}$) to denote tensors.
We denote the vectorization operator for 3D tensors (i.e., for reshaping a 3D array into a vector) as $\mathrm{vec}(\cdot): \mathbb{R}^{n_1 \times n_2 \times m} \rightarrow \mathbb{R}^{n}$. The vectorized tensor is $\mathbf{u} = \mathrm{vec}( \mathcal{U}) \in \mathbb{R}^{n}$, with $n = n_1 n_2 m$. Correspondingly, the inverse of the vectorization operator $\mathrm{vec}^{-1}(\cdot): \mathbb{R}^{n} \rightarrow \mathbb{R}^{n_1 \times n_2 \times m}$ denotes a tensorization operator. The relationship is summarized as follows:
\[
\mathcal{U} \in \mathbb{R}^{n_1 \times n_2 \times m}
\underset{\mathrm{vec}^{-1}}{\overset{\mathrm{vec}}{\rightleftharpoons}}
\mathbf{u} \in \mathbb{R}^{n}.
\]
The other major notations of the indices and variables that are used in this work are summarized in Table \ref{Tab:notation}. 
We denote the underlying signal or variable as $\tilde{\mathbf{u}}$, and its noisy measurement (resp. estimate) is denoted as $\mathbf{u}$ (resp. $\hat{\mathbf{u}}$).
The other notations used in our algorithms are discussed in later sections.

\subsection{Organization}

The rest of the paper is organized as follows. Section \ref{sec2} briefly discusses the recently proposed formulations for time-sequential signal denoising based on online and mini-batch sparsifying transform learning \cite{sai2015onlineTL,sai2015onlineTL2}. Then, Section \ref{sec3} presents the proposed online video processing framework, and two online approaches for denoising dynamic data. Section \ref{sec4} describes efficient algorithms for the proposed formulations. Section~\ref{sec5} demonstrates the behavior and promise of the proposed algorithms for denoising several datasets.
Section \ref{sec6} concludes with proposals for future work.

\begin{table}[t!]
\centering
\fontsize{9}{14pt}\selectfont
\begin{tabular}{|c|c|c|}
\hline
\textbf{Indices} & \textbf{Definition} & \textbf{Range} \\
\hline
$\tau$ & time index & $1,2,3,etc.$ \\
\hline
$p$ & spatial index of 3D patches in $\mathcal{Y}_{\tau}$ & $1...P$\\
\hline
$i$ & index of patches within mini-batch & $1...M$\\
\hline
$k$ & local mini-batches index at time $\tau$ & $1...N$\\
\hline
$j$ or $L_{k}^{\tau}$ & global mini-batch index & $1, 2,3, etc.$\\
\hline

\textbf{Variables} & \textbf{Definition} & \textbf{Dimension} \\
\hline
$\mathbf{W}_{\tau}$ & adaptive sparsifying transform & $n \times n$ \\
\hline
$\mathbf{Y}_{\tau} $ & video frames & $a \times b$\\
\hline
$\mathcal{Y}_{\tau} $ & input FIFO buffer & $a \times b \times m$\\
\hline
$\bar{\mathcal{Y}}_{\tau} $ & output FIFO buffer & $a \times b \times m$\\
\hline
$\mathbf{U}_j $ & mini-batch of vectorized data & $n \times M$\\
\hline
$\mathbf{X}_j $ & sparse codes of the mini-batch & $n \times M$\\
\hline
$\mathbf{v}_p$ & vectorized 3D patch & $n = n_1n_2m$\\
\hline

\textbf{Operators} & \textbf{Definition} & \textbf{Mapping} \\
\hline
$R_p$ &  extracts 3D patch in \textbf{A1} & $\mathbb{R}^{P} \rightarrow$ \\
\cline{1-2}
$B_p$ & forms 3D patch  in \textbf{A2} by BM & $\mathbb{R}^{n_1 \times n_2 \times m}$ \\
\hline
$R_p^{*}$ &  patch deposit operator in \textbf{A1} & $\mathbb{R}^{n_1 \times n_2 \times m}$ \\
\cline{1-2}
$B_p^{*}$ &  patch deposit operator in \textbf{A2} &  $ \rightarrow \mathbb{R}^{P}$\\
\hline

\end{tabular}
\caption{Notations of the indices and the main variables and operators.}
\label{Tab:notation}
\vspace{-0.2in}
\end{table}

\section{Signal Denoising via Online Transform Learning}  \label{sec2} 

The goal in denoising is to recover an estimate of a signal $\tilde{\mathbf{u}} \in \mathbb{R}^{n}$ from the measurement $\mathbf{u} = \tilde{\mathbf{u}} + \mathbf{e}$, corrupted by additive noise $\mathbf{e}$. 
Here, we consider a time sequence of noisy measurements $\left \{ \mathbf{u}_{t} \right \}$, with $\mathbf{u}_{t} = \tilde{\mathbf{u}}_{t} + \mathbf{e}_{t}$. 
We assume noise $\mathbf{e}_{t} \in \mathbb{R}^{n}$ whose entries are independent and identically distributed (i.i.d.) Gaussian with zero mean and possibly time-varying but known variance $\sigma_{t}^{2}$.  
Online denoising is to recover the estimates $\hat{\mathbf{u}}_{t}$ for $\tilde{\mathbf{u}}_{t}$ $\forall$ $t$ sequentially. 
Such time-sequential denoising with low memory requirements would be especially useful for streaming data applications.
We assume that the underlying signals $\left \{ \tilde{\mathbf{u}}_{t} \right \}$ are approximately sparse in an (unknown, or to be estimated) transform domain. 

\subsection{Online Transform Learning} 
In prior work \cite{sai2015onlineTL}, we proposed an online signal denoising methodology based on sparsifying transform learning, where the transform is adapted based on sequentially processed data.
For time $t = 1, 2, 3$, etc, the problem of updating the adaptive sparsifying transform and sparse code (i.e., the sparse representation in the adaptive transform domain) to account for the new noisy signal $\mathbf{u}_t  \in \mathbb{R}^{n}$ is
\begin{align} 
 \nonumber & \left\{\hat{\mathbf{W}}_{t}, \hat{\mathbf{x}}_{t}\right\} =  \underset{\mathbf{W},\,\mathbf{x}_{t}}{\arg \min}\:  \frac{1}{t} \sum _{\tau=1}^{t} \left\{ \left \| \mathbf{W} \mathbf{u}_{\tau} -\mathbf{x}_{\tau} \right \|_{2}^{2} + \lambda_{\tau} \nu(\mathbf{W})\right \}  \\
\nonumber & \;\;\;\; +  \frac{1}{t} \sum _{\tau=1}^{t} \alpha_{\tau}^{2} \left \| \mathbf{x}_{\tau} \right \|_{0} \,\,\,\,\,\,s.t.\; \:  \mathbf{x}_{\tau}= \hat{\mathbf{x}}_{\tau}, \; 1\leq \tau \leq t-1 \;\; (\mathrm{P1})
\end{align}
where the $\ell_0$ ``norm" counts the number of nonzeros in $x_{\tau}$, which is the sparse code of $\mathbf{u}_{\tau}$. 
Thus $\left \| \mathbf{W} \mathbf{u}_{\tau} -\mathbf{x}_{\tau} \right \|_{2}^{2}$ is the sparsification error (i.e., the modeling error in the transform model) for $\mathbf{u}_{\tau}$ in the transform $\mathbf{W}$. 
The term $\nu(\mathbf{W})= - \log \,\left | \mathrm{det \,} \mathbf{W} \right |  + \left \| \mathbf{W} \right \|_{F}^{2} $ is a transform learning regularizer \cite{sai2013tl}, 
$\lambda_{\tau} = \lambda_{0} \left \| \mathbf{u}_{\tau} \right \|_{2}^{2}$ with $\lambda_0 >0$ allows the regularizer term to scale with the first term in the cost, 
and the weight $\alpha_{\tau}$ is chosen proportional to  $\sigma_{\tau}$ (the standard deviation of noise in $\tilde{\mathbf{u}}_{\tau}$). 	
Matrix $\hat{\mathbf{W}}_{t}$ in (P1) is the optimal transform at time $t$, and $\hat{\mathbf{x}}_{t}$ is the optimal sparse code for $\mathbf{u}_t$.		

Note that at time $t$, only the latest optimal sparse code $\hat{\mathbf{x}}_{t}$ is updated in (P1)\footnote{This is because only the signal $\tilde{\mathbf{u}}_{t}$ is assumed to be stored in memory at time $t$ for the online scheme.} along with the transform $\hat{\mathbf{W}}_{t}$. 			%
The condition $\mathbf{x}_{\tau}= \hat{\mathbf{x}}_{\tau}, \; 1\leq \tau \leq t-1 $, is therefore assumed. 
For brevity, we will not explicitly restate this condition (or, its variants) in the formulations in the rest of this paper.
Although at each time $t$ the transform is updated based on all the past and present observed data, the online algorithm for (P1) \cite{sai2015onlineTL} involves efficient operations based on a few matrices of modest size, accumulated sequentially over time.

The regularizer $\nu(\mathbf{W})$ in (P1) prevents trivial solutions and controls the condition number and scaling of the learnt transform \cite{sai2013tl}. The condition number $\kappa(\mathbf{W})$ is upper bounded by a monotonically increasing function of $\nu(\mathbf{W})$ \cite{sai2013tl}.
In the limit $\lambda_{0} \to \infty$ (and assuming the $\mathbf{u}_{\tau}$, $1 \leq \tau \leq t$, are not all zero), the condition number of the optimal transform in (P1) tends to 1.
The specific choice of $\lambda_{0}$ (and hence the condition number) depends on the application.

\subsubsection{Denoising}
Given the optimal transform $\hat{\mathbf{W}}_{t}$ and the sparse code $\hat{\mathbf{x}}_{t}$, a simple estimate of the denoised signal is obtained as $\hat{\mathbf{u}}_{t} = \hat{\mathbf{W}}_{t}^{-1}\hat{\mathbf{x}}_{t}$.
Online transform learning can also be used for patch-based denoising of large images \cite{sai2015onlineTL}. 
Overlapping patches of the noisy images are processed sequentially (e.g., in raster scan order) via (P1), and the denoised image is obtained by averaging together the denoised patches at their respective image locations.

\subsubsection{Forgetting factor}
For non-stationary or highly dynamic data, it may not be desirable to uniformly fit a single transform $\mathbf{W}$ to all the $\mathbf{u}_{\tau}$, $1 \leq \tau \leq t$, in (P1).
Such data can be handled by introducing  a forgetting factor $\rho^{t-\tau}$ (with a constant $0<\rho<1$) that scales the terms in (P1)  \cite{sai2015onlineTL}. 
The forgetting factor diminishes the influence of ``old'' data. The objective function in this case is modified as
\begin{align} \label{forget}
\frac{1}{C_t}\sum _{\tau=1}^{t} \rho^{t-\tau}  \left\{ \left \| \mathbf{W} \mathbf{u}_{\tau}-\mathbf{x}_{\tau} \right \|_{2}^{2} + \lambda_{\tau} \nu(\mathbf{W}) + \alpha_{\tau}^{2} \left \| \mathbf{x}_{\tau} \right \|_{0}\right \}  \text{.}
\end{align}
where $C_t = \sum _{\tau=1}^{t} \rho^{t-\tau}$ is the normalization factor.

\subsection{Mini-batch learning} 
Another useful variation of Problem (P1) involves \emph{mini-batch} learning, where a block (group), or \textit{mini-batch} of signals is processed at a time \cite{sai2015onlineTL}.
Assuming a fixed mini-batch size $M$, the $L$th ($L \geq 1$) mini-batch of signals is $\mathbf{U}_{L} = \begin{bmatrix}\mathbf{u}_{LM - M + 1}\mid \mathbf{u}_{LM - M + 2}\mid &...& \mid \mathbf{u}_{LM}\end{bmatrix}$. For $L = 1, 2, 3$, etc, the mini-batch sparsifying transform learning problem is
\begin{align} 
\nonumber & (\mathrm{P2}) \;\; \left\{\hat{\mathbf{W}}_{L}, \hat{\mathbf{X}}_{L}\right\}   = \underset{\mathbf{W},\,\mathbf{X}_{L}}{\arg\min} \,  \frac{1}{LM} \sum _{j=1}^{L}  \left \| \mathbf{W} \mathbf{U}_{j}-\mathbf{X}_{j} \right \|_{F}^{2} \\
\nonumber & \;\;\;\;\;\;\;\;\;
 + \frac{1}{LM} \sum _{l=1}^{LM} \alpha_{l}^{2} \left \| \mathbf{x}_{l} \right \|_{0} + \frac{1}{LM} \sum _{j=1}^{L} \Lambda_{j} \, \nu(\mathbf{W})
\end{align} 
where the regularizer weight is $\Lambda_{j} =  \lambda_{0}  \begin{Vmatrix} \mathbf{U}_{j} \end{Vmatrix}_{F}^{2}$, and the matrix $\mathbf{X}_{L} =\; $ $ \begin{bmatrix}\mathbf{x}_{LM - M + 1}\mid \mathbf{x}_{LM - M + 2}\mid &...& \mid \mathbf{x}_{LM}\end{bmatrix}$ contains the block of sparse codes corresponding to $\mathbf{U}_{L}$. 

Since we only consider a finite number of frames or patches in practice (e.g., in the proposed VIDOSAT algorithms), the normalizations by $1/t$ in (P1), $1/C_{t}$ in (\ref{forget}), and $1/LM$ in (P2) correspondingly have no effect on the optimum $\begin{Bmatrix}\hat{\mathbf{W}}_t, \hat{\mathbf{X}}_{t} \end{Bmatrix}$ or $\begin{Bmatrix}\hat{\mathbf{W}}_L, \hat{\mathbf{X}}_{L} \end{Bmatrix}$. 
Thus we drop, for clarity\footnote{In practice, such normalizations may still be useful, to control the dynamic range of various internal variables in the algorithm.}, normalization factors from (P3) and all subsequent expressions for the cost functions. 

Once (P2) is solved, a simple denoised estimate of the noisy block of signals in $\mathbf{U}_L$ is obtained as $\hat{\mathbf{U}}_{L} = \hat{\mathbf{W}}_{L}^{-1}\hat{\mathbf{X}}_{L}$.
The mini-batch transform learning Problem (P2) is a generalized version of (P1), with (P2) being equivalent to (P1) for $M=1$.
Similar to (\ref{forget}), (P2) can be modified to include a forgetting factor. 
Mini-batch learning can provide potential speedups over the $M=1$ case in applications, but this comes at the cost of higher memory requirements and latency (i.e., delay in producing output) \cite{sai2015onlineTL}.

\section{VIDOSAT Framework and Formulations}  \label{sec3}

Prior work on adaptive sparsifying transform-based image denoising \cite{sai2015closed,wen2015octobos,sai2015onlineTL} adapted the transform operator to 2D image patches. However, in video denoising, exploiting the sparsity and redundancy in both the spatial and temporal dimensions typically leads to better performance than denoising each frame separately \cite{elad2010sksvd}. We therefore propose an online approach to video denoising by learning a sparsifying transform on appropriate 3D spatio-temporal patches.

\subsection{Video Streaming and Denoising Framework} \label{sec31}
Fig. \ref{frame} illustrates the framework of our proposed online denoising scheme for streaming videos. 
The frames of the noisy video (assumed to be corrupted by additive i.i.d. Gaussian noise) denoted as $\mathbf{Y}_{\tau} \in \mathbb{R}^{a \times b}$ arrive at $\tau = 1, 2, 3,$ etc. 
At time $\tau = t$, the newly arrived frame $\mathbf{Y}_{t}$ is added to a fixed-size FIFO (first in first out) buffer (i.e., queue) that stores a block of $m$ consecutive frames $\begin{Bmatrix} \mathbf{Y}_{i} \end{Bmatrix}_{i=t-m+1}^{t}$. 
The oldest (leftmost) frame is dropped from the buffer at each time instant.
We denote the spatio-temporal tensor or 3D array obtained by stacking noisy frames along the temporal dimension as $\mathcal{Y}_{t} = \begin{bmatrix} \mathbf{Y}_{t - m + 1} \mid &...& \mid \mathbf{Y}_{t}\end{bmatrix} \in \mathbb{R}^{a\times b \times m}$. 
We denoise the noisy array $\mathcal{Y}_{t}$ using the proposed VIDOSAT mini-batch denoising algorithms (denoted by the red box in Fig. \ref{frame}) that are discussed in Sections \ref{sec32} and \ref{sec4}.
These algorithms denoise groups (mini-batches) of 3D patches sequentially and adaptively, by learning sparsifying transforms.
Overlapping patches are used in our framework. 

The patches output by the mini-batch denoising algorithms are deposited at their corresponding spatio-temporal locations in the
fixed-size FIFO output $\bar{\mathcal{Y}}_{t} = \begin{bmatrix} \bar{\mathbf{Y}}_{t - m + 1} \mid &...& \mid \bar{\mathbf{Y}}_{t}\end{bmatrix}$ by adding them to the contents of $\bar{\mathcal{Y}}_t$. 
We call this process \emph{patch aggregation}.
The streaming scheme then outputs the oldest frame $\bar{\mathbf{Y}}_{t - m + 1}$. 
The denoised estimate $\hat{\mathbf{Y}}_{t - m + 1}$ is obtained by normalizing $\bar{\mathbf{Y}}_{t - m + 1}$ pixel-wise by the number of occurrences of each pixel in the aggregated patches. (see Section \ref{sec4} for details). 

\begin{figure}[!t]
\centering
\includegraphics[width=3.3in]{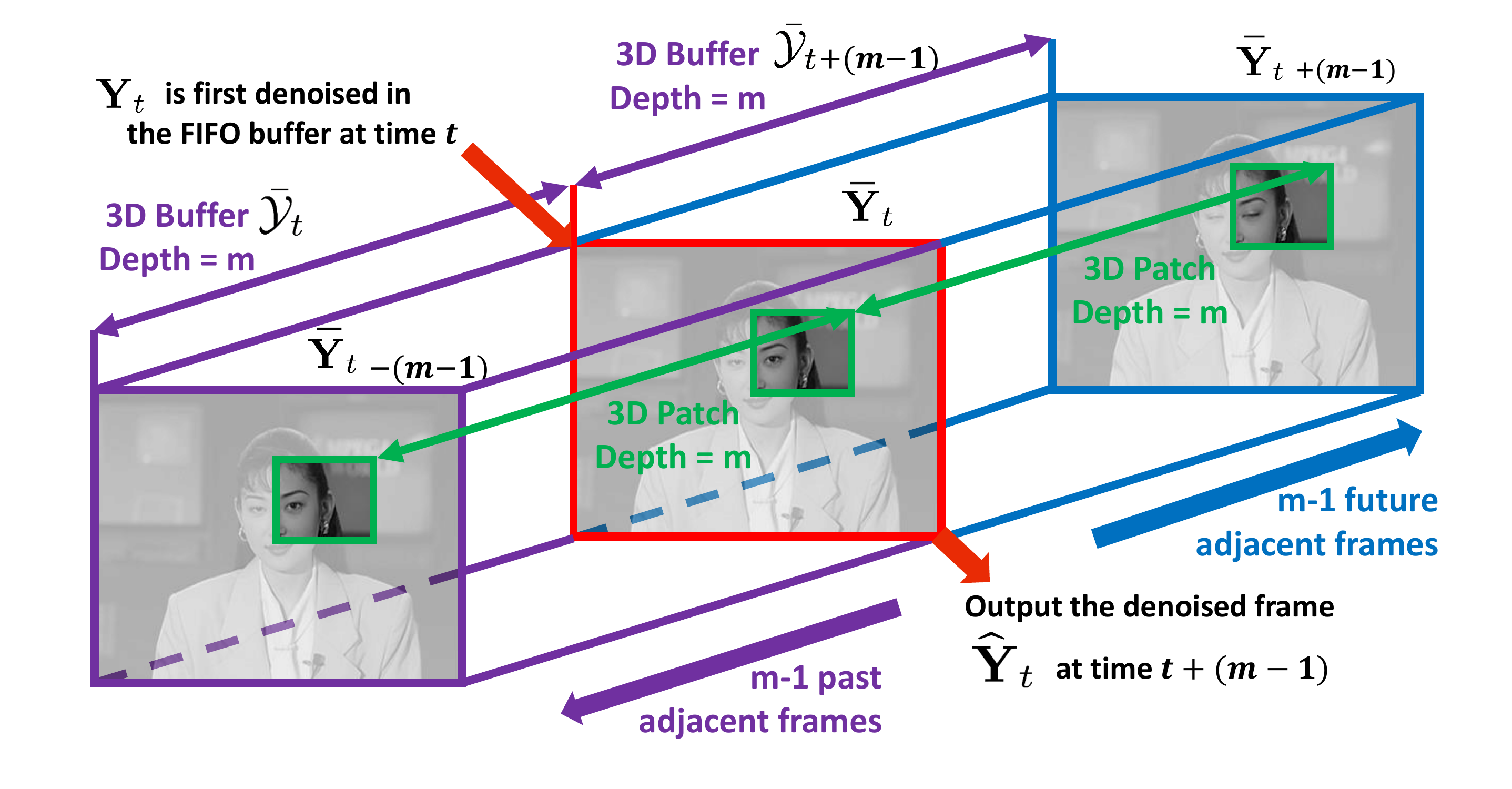}
\vspace{-0.1in}
\caption{Illustration of the output buffer from time $t$ to $t+(m-1)$ for generating the denoised frame output $\hat{\mathbf{Y}}_t$.}
\label{fig:buffer}
\vspace{-0.1in}
\end{figure}

Though any frame could be denoised and output from $\bar{\mathcal{Y}}_{t}$ instantaneously, we observe improved denoising quality by averaging over multiple denoised estimates at different time.  
Fig. \ref{fig:buffer} illustrates how the output buffer varies from time $t$ to $t+(m-1)$, to output the denoised $\hat{\mathbf{Y}}_t$.
In practice, we set the length of the output buffer $\bar{\mathcal{Y}}$ to be the same as the 3D patch depth $m$, such that each denoised frame $\hat{\mathbf{Y}}_t$ is output by averaging over its estimates from all 3D patches that group the $t$th frame with $m-1$ adjacent frames. 
We refer to this scheme as ``two-sided" denoising, since the $t$th frame is denoised together with both past and future adjacent frames  ($m-1$ frames on each side), which are highly correlated.
Now, data from frame $\mathbf{Y}_t$ is contained in 3D patches that also contain data from frame $\mathbf{Y}_{t+m-1}$. 
Once these patches are denoised, they will contribute (by aggregation into the output buffer) to the final denoised frame $\hat{\mathbf{Y}}_t$. 
Therefore, we must wait for frame $\mathbf{Y}_{t+m-1}$ before producing the final estimate $\hat{\mathbf{Y}}_t$.
Thus there is a delay of $m - 1$ frames between the arrival of the noisy $\mathbf{Y}_{t}$ and the generation of its final denoised estimate $\hat{\mathbf{Y}}_{t}$. 

\subsection{VIDOSAT Mini-Batch Denoising Formulation} \label{sec32}

Here, we discuss the mini-batch denoising formulation that is a core part
of the proposed online video denoising framework. 
For each time instant $t$, we denoise $P$ partially overlapping size $n_1 \times n_2 \times m$ 3D patches of $\mathcal{Y}_t$ whose vectorized versions are denoted as $\begin{Bmatrix} \mathbf{v}_{p}^{t} \end{Bmatrix}_{p=1}^{P}$, with $\mathbf{v}_{p}^{t} \in \mathbb{R}^{n}$, $n=m n_1n_2$. 
We sequentially process disjoint groups 
of $M$ such patches, and
the groups or mini-batches of patches (total of $N$ mini-batches, where $P = MN$) are denoted as
$\begin{Bmatrix} \mathbf{U}_{L_{k}^{t}} \end{Bmatrix}_{k=1}^{N}$, with $\mathbf{U}_{L_{k}^{t}} \in \mathbb{R}^{n \times M}$. 
Here, $k$ is the \emph{local} mini-batch index within the set of $P$ patches of $\mathcal{Y}_t$, 
whereas $L_{k}^{t} \triangleq N \times (t-1) + k$ is the \emph{global} mini-batch index, identifiying the mini-batch in both time $t$ and location within the set of $P$ patches of $\mathcal{Y}_t$.

For each $t$, we solve the following online transform learning problem for each $k = 1, 2, 3,..., N$, to adapt the transform and sparse codes 
sequentially to the mini-batches in $\mathcal{Y}_{t}$:
\begin{align} 
\nonumber (\mathrm{P3}) \;\;  & \left\{\hat{\mathbf{W}}_{L_{k}^{t}}, \hat{\mathbf{X}}_{L_{k}^{t}}\right\}  = \,\underset{\mathbf{W},\mathbf{X}_{L_{k}^{t}}}{\arg\min}  \sum _{j=1}^{L_{k}^{t}} \rho^{L_{k}^{t}-j} \left \| \mathbf{W} \mathbf{U}_{j} - \mathbf{X}_{j} \right \|_{F}^{2} \\
\nonumber & + \sum _{j=1}^{L_{k}^{t}} \rho^{L_{k}^{t}-j} \left\{
\Lambda_{j} \, \nu (\mathbf{W}) + \sum_{i=1}^{M} \alpha_{j, i}^{2} \begin{Vmatrix} \mathbf{x}_{j, i} \end{Vmatrix}_{0}\right\}. 
\end{align}
Here, the transform is adapted based on patches from \emph{all} the \emph{observed} $\mathbf{Y}_{\tau}$, $1 \leq \tau \leq t$. 
The matrix $\mathbf{X}_{j} =  \begin{bmatrix} \mathbf{x}_{j, 1} \mid ... \mid \mathbf{x}_{j, M}\end{bmatrix} \in \mathbb{R}^{n \times M}$ denotes the transform sparse codes corresponding to the mini-batch $\mathbf{U}_j$. 
The sparsity penalty weight $\alpha_{j,i}^2$ in (P3) controls the number of non-zeros in $\mathbf{x}_{j,i}$.
We set $\alpha_{j, i} = \alpha_0 \sigma_{j, i}$, where $\alpha_0$ is a constant and $\sigma_{j, i}$ is the noise standard deviation for each patch.
We use a forgetting factor $\rho^{L_{k}^{t}-j}$ in (P3) to diminish the influence of old frames and old mini-batches.

Once (P3) is solved, the denoised version of the current noisy mini-batch $\hat{\mathbf{U}}_{L_{k}^{t}}$ is computed. The columns of the denoised $\hat{\mathbf{U}}_{L_{k}^{t}}$ are tensorized and aggregated 
at the corresponding spatial and temporal locations in the output FIFO buffer.
Section \ref{sec4} 
next discusses the proposed VIDOSAT algorithms in full detail.


\section{Video Denoising Algorithms}  \label{sec4}

We now discuss 
two video denoising algorithms,
namely VIDOSAT and VIDOSAT-BM. 
VIDOSAT-BM uses block matching to generate the 3D patches from $\mathcal{Y}_t$.
Though these methods differ in the way they construct the 3D patches, and the way the denoised patches are aggregated in the output FIFO,
they both denoise groups of 3D patches sequentially by solving (P3). 
The VIDOSAT denoising algorithm (without BM) is summarized in Algorithm $\mathbf{A1}$\footnote{In practice, we wait for the first $m$ frames to be received, before starting Algorithm $\mathbf{A1}$, to avoid zero frames in the input FIFO buffer}.
The VIDOSAT-BM algorithm, a modified version of Algorithm $\mathbf{A1}$, is discussed in Section \ref{sec42}.

\begin{figure}
\begin{tabular}{p{8.3cm}}
\hline
Algorithm \textbf{A1}: VIDOSAT Denoising Algorithm \\
\hline
\vspace{-0.05in}
\textbf{Input:} The noisy frames $\mathbf{Y}_{\tau}$ ($\tau=1,2,3,$ etc.), and the initial transform $\mathbf{W}_{0}$ (e.g., 3D DCT). \\
\textbf{Initialize:} $\hat{\mathbf{W}} = \mathbf{W}_{0}$, $\mathbf{\Gamma} = \mathbf{\Theta} = \mathbf{0}$, $\beta = 0$,\\ and output buffer $\bar{\mathcal{Y}} = 0$. \\
\textbf{For $\;\tau = 1, 2, 3,$ etc., Repeat}\\
The newly arrived frame $\mathbf{Y}_{\tau} \rightarrow $ latest frame in the input FIFO frame buffer $\mathcal{Y}$. \\
\textbf{For $\;k = 1, ..., N$ Repeat}\\
~~~Indices of patches in $\mathcal{Y}$:~~~ $S_{k} = \{M(k-1) + 1, ..., Mk \}$.\\
\begin{enumerate}						

\item \textbf{Noisy Mini-Batch Formation:}	
\begin{enumerate}	
\item Patch Extraction:~~$\mathbf{v}_{p} = \mathrm{vec}(R_p \mathcal{Y})$ ~ $\forall p \in S_{k}$.
\item $\mathbf{U} = \begin{bmatrix} \mathbf{u}_1 \mid ... \mid \mathbf{u}_M \end{bmatrix} \leftarrow \begin{bmatrix} \mathbf{v}_{Mk-M+1} \mid ... \mid \mathbf{v}_{Mk} \end{bmatrix}$. 
\end{enumerate}

\item \textbf{Sparse Coding:}
$\hat{\mathbf{x}}_{i} =  H_{\alpha_{i}} (\hat{\mathbf{W}} \mathbf{u}_{i})$ ~$\forall i \in \left \{ 1, ..., M \right \}$.

\item \textbf{Mini-batch Transform Update:} 
\begin{enumerate}	
\item Define $\Lambda \triangleq \lambda_0 \| \mathbf{U} \|_{F}^{2}$ and $\hat{\mathbf{X}} \triangleq \begin{bmatrix} \hat{\mathbf{x}}_1 \mid ... \mid \hat{\mathbf{x}}_M \end{bmatrix}$.
\vspace{0.03in}
\item $\mathbf{\Gamma} \leftarrow \rho \mathbf{\Gamma} + \mathbf{U} \mathbf{U}^{T}$.

\item $\mathbf{\Theta} \leftarrow \rho \mathbf{\Theta} + \mathbf{U} \hat{\mathbf{X}}^T$.

\item $\beta \leftarrow \rho \beta + \Lambda$.

\item Matrix square root: $\mathbf{Q} \leftarrow (\mathbf{\Gamma} + \beta \mathbf{I})^{1/2}$.

\item Full SVD: $\mathbf{\Phi} \mathbf{\Sigma} \mathbf{\Psi}^{T} \leftarrow $ SVD$(\mathbf{Q}^{-1} \mathbf{\Theta})$.
\item $\hat{\mathbf{W}} \leftarrow 0.5 \mathbf{\Psi} \left(\mathbf{\Sigma}+ \left ( \mathbf{\Sigma}^{2}+2\beta \mathbf{I} \right )^{\frac{1}{2}}\right)\mathbf{\Phi}^{T}\mathbf{Q}^{-1}$.
\end{enumerate}

\item \textbf{3D Denoised Patch Reconstruction:} 
\begin{enumerate}	
\item Update Sparse Codes: $\hat{\mathbf{x}}_{i} =  H_{\alpha_{i}} (\hat{\mathbf{W}} \mathbf{u}_{i})$~ $\forall i$.

\item Denoised mini-batch: $\hat{\mathbf{U}} = \hat{\mathbf{W}}^{-1} \hat{\mathbf{X}}$.
\item $\begin{bmatrix} \hat{\mathbf{v}}_{M(k-1)+1} \mid ... \mid \hat{\mathbf{v}}_{Mk} \end{bmatrix} \leftarrow \hat{\mathbf{U}}$
\item Tensorization: $\hat{\mathcal{V}}_{p} = \mathrm{vec}^{-1}(\hat{\mathbf{v}}_{p})$ $\forall p \in S_{k}$.
\end{enumerate}

\item \textbf{Aggregation:} Aggregate patches $\begin{Bmatrix} \hat{\mathcal{V}}_p \end{Bmatrix}$ at corresponding locations: $\bar{\mathcal{Y}} \leftarrow \sum_{p \in S_k} R_p^* \hat{\mathcal{V}}_p$.

\end{enumerate}
\textbf{End}\\
\textbf{Output:} The oldest frame in $\bar{\mathcal{Y}}$ after normalization $\rightarrow$ the denoised frame $\hat{\mathbf{Y}}_{\tau - m+1}$.\\
\textbf{End}\\

\hline
\end{tabular} \label{A1}
\end{figure}


\subsection{VIDOSAT}  \label{sec41}

As discussed in Section \ref{sec32}, the VIDOSAT algorithm processes each mini-batch $\mathbf{U}_j $ in $\mathcal{Y}_\tau$ sequentially.
We solve the mini-batch transform learning problem (P3) using a simple alternating minimization approach, with one alternation per mini-batch, which works well and saves computation.
Initialized with the most recently estimated transform (warm start), we perform two steps for (P3): Sparse Coding, and Mini-batch Transform Update, which compute $\hat{\mathbf{X}}_{j}$ and update $\hat{\mathbf{W}}_{j}$, respectively. 
Then, we compute the denoised mini-batch $\hat{\mathbf{U}}_{j}$, and 
aggregate the denoised patches into the output buffer $\bar{\mathcal{Y}}_{\tau}$. 

The major steps of the VIDOSAT algorithm \textbf{A1} for denoising the $k$th mini-batch $\mathbf{U}_{L_{k}^{t}}$ at time $t$ and further processing these denoised patches are described below. 
To facilitate the exposition and interpretation in terms of the general online denoising algorithm described, various quantities (such as positions of 3D patches in the video stream) are indexed in the text with respect to absolute time $t$. 
On the other hand, to emphasize the streaming nature of Algorithm \textbf{A1} and its finite (and modest) memory requirements, indexing of internal variables in the statement of the algorithm is local.

\subsubsection{Noisy Mini-Batch Formation}
To construct each mini-batch $\mathbf{U}_{L_{k}^{t}}$, partially overlapping size $n_{1} \times n_{2} \times m$ 3D patches of $\mathcal{Y}_t$ are extracted sequentially in a spatially contiguous order (raster scan order with direction reversal on each line)\footnote{We did not observe any marked improvement in denoising performance, when using other scan orders such as raster or Peano-Hilbert scan \cite{ouni2012scan}.}. 
Let $R_{p} \mathcal{Y}_t$
denote the $p$th vectorized 3D patch of $\mathcal{Y}_t$, with $R_p$ being the
patch-extraction operator. 
Considering the patch indices $S_{k} = \begin{Bmatrix} M(k-1) + 1, ..., Mk \end{Bmatrix}$ for the $k$th mini-batch, 
we extract $\begin{Bmatrix} \mathbf{v}_p^t = \mathrm{vec}(R_{p} \mathcal{Y}_t) \end{Bmatrix}_{p \in S_{k}}$ as the patches in the mini-batch.
Thus $\mathbf{U}_{L_{k}^{t}} = \begin{bmatrix} \mathbf{v}_{M(k-1)+1}^{t} \mid ... \mid \mathbf{v}_{Mk}^{t} \end{bmatrix}$.
To impose spatio-temporal contiguity of 3D patches extracted from two adjacent stacks of frames, we reverse the raster scan order (of patches) between $\mathcal{Y}_t$ and $\mathcal{Y}_{t+1}$.

\subsubsection{Sparse Coding}
Given the sparsifying transform $\mathbf{W} = \hat{\mathbf{W}}_{{L_{k}^{t}}-1}$ estimated for the most recent mini-batch, we solve Problem (P3) for the sparse coefficients $\hat{\mathbf{X}}_{L_{k}^{t}}$:
\begin{equation} \label{sparse1}
\hat{\mathbf{X}}_{L_{k}^{t}} =\underset{\mathbf{X}}{\arg\min} \begin{Vmatrix} \mathbf{W} \mathbf{U}_{L_{k}^{t}} - \mathbf{X} \end{Vmatrix}_{F}^{2} + \sum_{i=1}^{M} \alpha_{L_{k}^{t}, i}^{2} \begin{Vmatrix} \mathbf{x}_i \end{Vmatrix}_{0}
\end{equation}
A solution for (\ref{sparse1}) is given in closed-form as $\hat{\mathbf{x}}_{L_{k}^{t}, i} =  H_{\alpha_{L_{k}^{t}, i}} (\hat{\mathbf{W}}_{L_{k}^{t}} \mathbf{u}_{L_{k}^{t}, i})$ $\forall$ $i$ \cite{sai2015onlineTL}. Here, the hard thresholding operator $H_{\alpha} (\cdot):\, \mathbb{R}^n \rightarrow \mathbb{R}^n$ is applied to a vector element-wise, as defined by
\begin{equation} \label{equ88}
 \left ( H_{\alpha} (\mathbf{d}) \right )_{r}=\left\{\begin{matrix}
 0&, \;\;\left | d_{r} \right | < \alpha \\
d_{r}  & ,\;\;\left | d_{r} \right | \geq \alpha 
\end{matrix}\right.
\end{equation}
This simple hard thresholding operation for transform sparse coding is similar to traditional techniques involving analytical sparsifying transforms \cite{mallat1999wavelet}. 

\subsubsection{Mini-batch Transform Update} 
We solve Problem (P3) for $\mathbf{W}$ with fixed $\mathbf{X}_{j} = \hat{\mathbf{X}}_j$, $1\leq j \leq L_{k}^{t}$, as follows:
\begin{equation}  \label{tru2}
\underset{\mathbf{W}}{\min}\: \sum _{j=1}^{L_{k}^{t}}   \rho^{L_{k}^{t} -j}  \left \{ \begin{Vmatrix} \mathbf{W} \mathbf{U}_{j}- \mathbf{X}_{j} \end{Vmatrix}_{F}^{2} + \Lambda_{j} \nu(\mathbf{W}) \right \}.
\end{equation}
This problem has a simple solution (similar to Section III-B2 in \cite{sai2015onlineTL}). 
Set index $J=L_t^k$, and define the following quantities:
$\mathbf{\Gamma}_J \triangleq \sum_{j=1}^{J} \rho^{J -j} \mathbf{U}_{j}  \mathbf{U}_{j}^{T}$, 
$\mathbf{\Theta}_J \triangleq \sum_{j=1}^J \rho^{J-j}  {\mathbf{U}}_j \hat{\mathbf{X}}_j^T $, 
and 
$\beta_J \triangleq \sum_{j=1}^J \rho^{J-j} \Lambda_j$.
Let $\mathbf{Q} \in \mathbb{R}^{n \times n}$ be a square root  (e.g., Cholesky factor) of $(\mathbf{\Gamma}_J + \beta_J \mathbf{I})$, i.e.,  $\mathbf{Q} \mathbf{Q}^T = \mathbf{\Gamma}_J + \beta_J \mathbf{I}$. 
Denoting the full singular value decomposition (SVD) of $\mathbf{Q}^{-1} \mathbf{\Theta}_J$ as $\mathbf{\Phi} \mathbf{\Sigma} \mathbf{\Psi}^{T}$, we then have that the closed-form solution to (\ref{tru2}) is
\begin{equation} \label{tru3b}
\hat{\mathbf{W}}_{J}= 0.5 \mathbf{\Psi} \left(\mathbf{\Sigma}+ \left ( \mathbf{\Sigma}^{2}+2\beta_J \mathbf{I} \right )^{\frac{1}{2}}\right)\mathbf{\Phi}^{T}\mathbf{Q}^{-1}
\end{equation} 
where $\mathbf{I}$ denotes the identity matrix, and $(\cdot)^{\frac{1}{2}}$ denotes the positive definite square root of a positive definite (diagonal) matrix. 
The quantities $\mathbf{\Gamma}_{J}$, 
$\mathbf{\Theta}_{J}$, 
and $\beta_{J}$ are all computed sequentially over time $t$ and mini-batches $k$ \cite{sai2015onlineTL}.

\subsubsection{3D Denoised Patch Reconstruction} 
We denoise $ {\mathbf{U}}_{L_{k}^{t}}$ using the updated transform. 
First, we repeat the sparse coding step using the updated $\hat{\mathbf{W}}_{L_{k}^{t}}$ as $\hat{\mathbf{x}}_{L_{k}^{t}, i} =  H_{\alpha_{L_{k}^{t}, i}} (\hat{\mathbf{W}}_{L_{k}^{t}}  {\mathbf{u}}_{L_{k}^{t}, i})$ $\forall$ $i$. 
Then, with fixed $\hat{\mathbf{W}}_{L_{k}^{t}}$ and $\hat{\mathbf{X}}_{L_{k}^{t}}$, the denoised mini-batch is obtained in the least squares sense under the transform model as 
\begin{equation} \label{recon}
\hat{\mathbf{U}}_{L_{k}^{t}} = \hat{\mathbf{W}}_{L_{k}^{t}}^{-1}\hat{\mathbf{X}}_{L_{k}^{t}}.
\end{equation} 
The denoised mini-batch is used to update the denoised (vectorized) 3D patches as
$\hat{\mathbf{v}}_{M(k-1)+i}^{t} = \hat{\mathbf{u}}_{L_{k}^{t}, i}$ $\forall i$. All reconstructed vectors $\begin{Bmatrix} \hat{\mathbf{v}}_{p}^{t} \end{Bmatrix}_{p \in S_k}$ from the $k$th mini-batch denoising result are tensorized as 
$\begin{Bmatrix} \mathrm{vec}^{-1}(\hat{\mathbf{v}}_{p}^{t}) \end{Bmatrix}_{p \in S_k}$.

\begin{figure}[!t]
\centering
\includegraphics[width=3.3in]{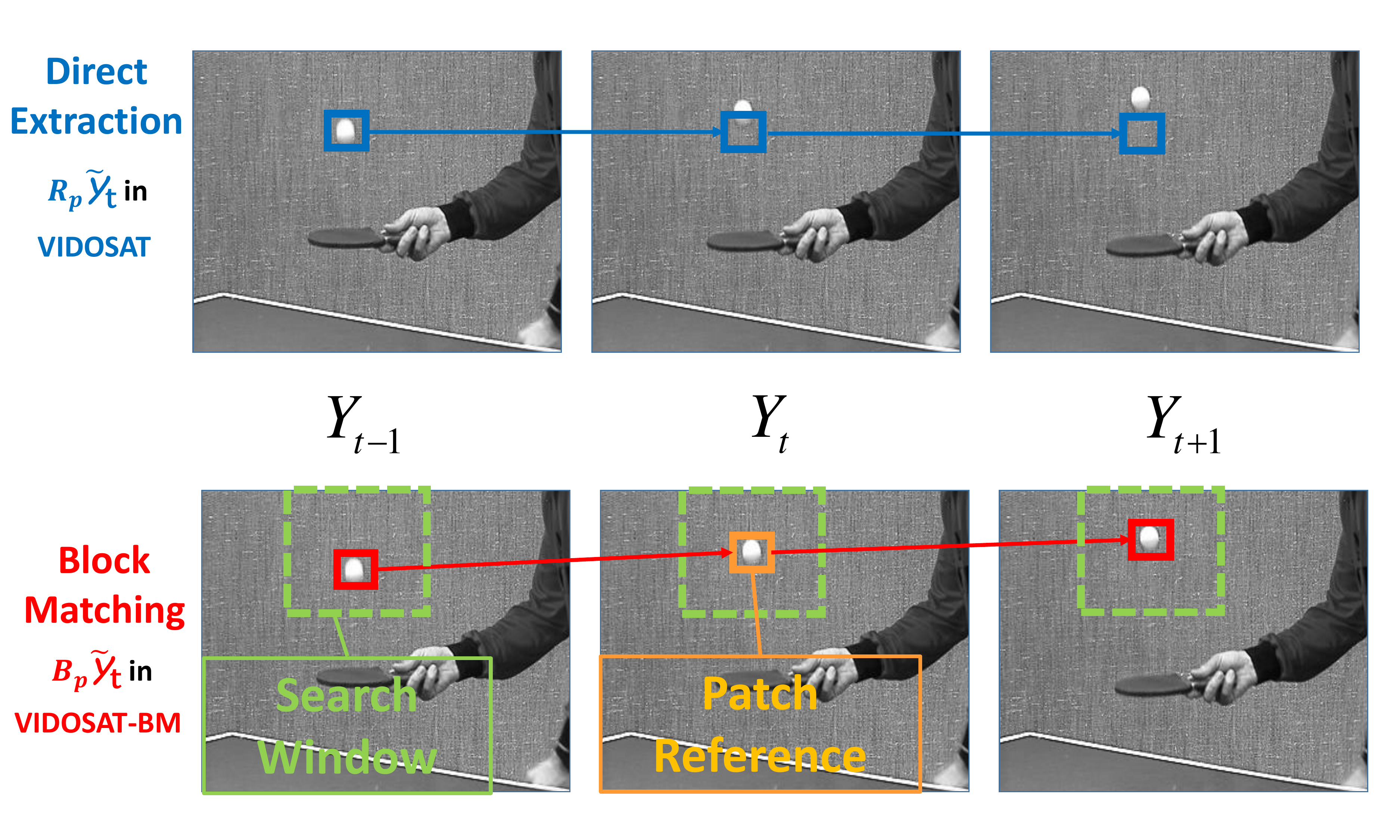}
\vspace{-0.1in}
\caption{Illustration of the different 3D patch construction methods in VIDOSAT (blue) and VIDOSAT-BM (red). The 3D search window used in VIDOSAT-BM is illustrated in green.}
\label{fig:BM}
\vspace{-0.1in}
\end{figure}

\subsubsection{Aggregation} 

\begin{figure}[!t]
\centering
\includegraphics[width=3.3in]{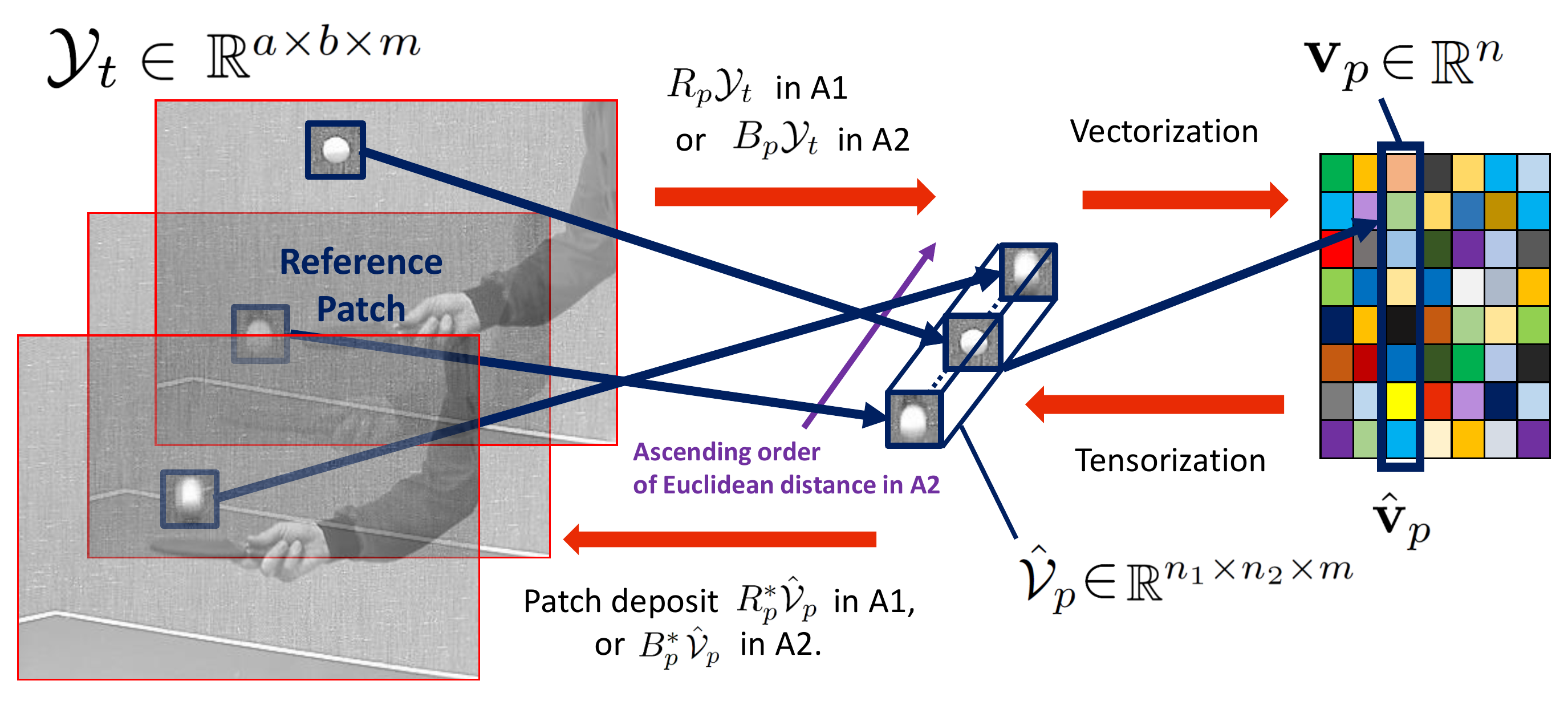}
\vspace{-0.1in}
\caption{Patch deposit $R_p^* \, \mathrm{vec}^{-1}(\hat{\mathbf{v}}_{p})$ (resp. $B_p^* \, \mathrm{vec}^{-1}(\hat{\mathbf{v}}_{p})$)  as an adjoint of patch extraction operator in \textbf{A1} (resp. an adjoint of BM operator in \textbf{A2}).}
\label{fig:deposit}
\vspace{-0.1in}
\end{figure}

The denoised 3D patches 
$\begin{Bmatrix} \mathrm{vec}^{-1}(\hat{\mathbf{v}}_{p}^{t}) \end{Bmatrix}_{p \in S_k}$ 
from each mini-batch are sequentially aggregated at their corresponding spatial and temporal locations in the output FIFO buffer as
$\sum_{p \in S_k} R_p^* \, \mathrm{vec}^{-1}(\hat{\mathbf{v}}_{p}^{t}) \rightarrow \bar{\mathcal{Y}}_{t} \in \mathbb{R}^{a \times b \times m}$, where the adjoint $R_p^*$ is the patch deposit operator. Fig. \ref{fig:deposit} illustrates the patch deposit procedure for aggregation.

When all $N$ denoised mini-batches for $ {\mathcal{Y}}_t$ are generated, 
and the patch aggregation in $\bar{\mathcal{Y}}_{t}$ completes, 
the oldest frame in $\bar{\mathcal{Y}}_{t}$ is normalized pixel-wise by the number of occurrences (which ranges from $2m - 1$, for pixels at the corners of a video frame, to $n$ for pixels away from the borders of a video frame) of that pixel among patches aggregated into the output buffer. 
This normalized result is output as the denoised frame $\hat{\mathbf{Y}}_{t-m+1}$.

\subsection{VIDOSAT-BM} \label{sec42}

For videos with relatively static scenes, each extracted spatio-temporal tensor $R_{p} \mathcal{Y}_t$ in the VIDOSAT Algorithm \textbf{A1}
typically has high temporal correlation, 
implying high (3D) transform domain sparsity.
However, highly dynamic videos usually involve various motions, such as translation, rotation, scaling, etc. 
Figure \ref{fig:BM} demonstrates one example when the 3D patch construction strategy in the VIDOSAT denoising algorithm \textbf{A1} fails to capture 
the properties of the moving object.
Thus, Algorithm \textbf{A1} could provide sub-optimal denoising performance for highly dynamic videos.
We propose an alternative algorithm, dubbed VIDOSAT-BM, which improves VIDOSAT denoising by constructing 3D patches using block matching. 

The proposed VIDOSAT-BM solves the online transform learning problem (P3) with 
 a different methodology for constructing the 3D patches and each mini-batch.
The Steps $(2)-(4)$ in Algorithm \textbf{A1} remain the same for VIDOSAT-BM. 
We now discuss the modified Steps $(1)$ and $(5)$ in the VIDOSAT-BM denoising algorithm, to which we also refer as Algorithm \textbf{A2}.

\textbf{3D Patch and Mini-Batch Formation in VIDOSAT-BM}: Here, we use a small and odd-valued sliding (temporal) window size $m$ (e.g., we set $m=9$ in the video denoising experiments in Section \ref{sec5}, which corresponds to $\sim 0.2$s buffer duration for a video with $40$ Hz frame rate). Within the $m$-frame input FIFO buffer $\mathcal{Y}_t$, we approximate the various motions in the video using simple (local) translations \cite{le1991mpeg}.

We consider the middle frame $\mathbf{Y}_{t-(m-1)/2}$ in the input FIFO buffer $\mathcal{Y}_t$, and sequentially extract all 2D overlapping patches $ {\mathbf{Z}}_{p}^{t} \in \mathbb{R}^{n_1 \times n_2}$, $1 \leq p \leq P$ in $\mathbf{Y}_{t-(m-1)/2}$, in a 2D spatially contiguous (raster scan) order. 
For each $ {\mathbf{Z}}_{p}^{t}$, we form a $h_1 \times h_2 \times m$ pixel local search window centered at the center of $ {\mathbf{Z}}_{p}^{t}$ (see the illustration in Fig. \ref{fig:BM}). 
We apply a spatial BM operator, denoted $B_p$, to find (using exhaustive search) the $(m-1)$ patches, one for each neighboring frame in the search window, that are most similar to $ {\mathbf{Z}}_{p}^{t}$ in Euclidean distance.
The operator $B_p$ stacks the $ {\mathbf{Z}}_{p}^{t}$, followed by the $(m-1)$ matched patches, in an ascending order of their Euclidean distance to $ {\mathbf{Z}}_{p}^{t}$, to form the $p$th 3D patch $B_{p} {\mathcal{Y}}_t \in \mathbb{R}^{n_1 \times n_2 \times m}$.
Similar BM approaches have been used in prior works on video compression (e.g., MPEG) for motion compensation \cite{le1991mpeg}, and in recent works on spatiotemporal medical imaging \cite{yoon2014motion}.
The coordinates of all selected 2D patches are recorded to be used later in the denoised patch aggregation step. 
Instead of 
constructing the 3D patches from 2D patches in corresponding locations in contiguous frames (i.e., $R_p  {\mathcal{Y}}_t$ in Algorithm A1),
we form the patches using BM and work with the vectorized $ {\mathbf{v}}_{p}^{t} = \mathrm{vec}(B_{p} {\mathcal{Y}}_t) \in \mathbb{R}^{n}$ in VIDOSAT-BM.
The $k$-th mini-batch is defined as in Algorithm \textbf{A1} as $  {\mathbf{U}}_{L_{k}^{t}} = \begin{bmatrix}  {\mathbf{v}}_{M(k-1)+1}^{t} \mid ... \mid  {\mathbf{v}}_{Mk}^{t} \end{bmatrix}$.

\textbf{Aggregation}: 
Each denoised 3D patch (tensor) of $\begin{Bmatrix} \mathrm{vec}^{-1}(\hat{\mathbf{v}}_{p}^{t}) \end{Bmatrix}_{p \in S_k}$ contains the matched (and denoised) 2D patches.
They are are sequentially aggregated
at their recorded spatial and temporal locations in the output FIFO buffer $\bar{\mathcal{Y}}_{t}$ as
$\sum_{p \in S_k} B_p^* \, \mathrm{vec}^{-1}(\hat{\mathbf{v}}_{p}^{t}) \rightarrow \bar{\mathcal{Y}}_{t} \in \mathbb{R}^{a \times b \times m}$, where the adjoint $B_p^*$ is the patch deposit operator in \textbf{A2}. Fig. \ref{fig:deposit} illustrates the patch deposit procedure for aggregation in \textbf{A2}.
Once the aggregation of $\bar{\mathcal{Y}}_{t}$ completes, the oldest frame in $\bar{\mathcal{Y}}_{t}$ is normalized pixel-wise by the number of occurrences of each pixel among patches in the denoising algorithm. 
Unlike Algorithm \textbf{A1} where this number of occurrences is the same for all frames, in Algorithm \textbf{A2} this number is data-dependent and varies from frame to frame and pixel to pixel. We record the number of occurrences of each pixel which is based on the recorded locations of the matched patches, and can be computed online as described.
The normalized oldest frame is output by Algorithm \textbf{A2} for each time instant.

\vspace{-0.1in}

\begin{table*}[t]
\centering
\fontsize{9}{16pt}\selectfont
\begin{tabular}{|c|c|c|c|c|c|c|c|c|c|c|c|c|}
\hline
Data &      \multicolumn{5}{c|}{\textbf{ASU} Dataset ($26$ videos)} & $\Delta$PSNR &
\multicolumn{5}{c|}{\textbf{LASIP} Dataset ($8$ videos)} & $\Delta$PSNR \\
\cline{1-6} \cline{8-12}
$\sigma$ & 5 & 10 & 15 & 20 & 50 & (std.) &
5 & 10 & 15 & 20 & 50 & (std.)  \\
\hline
fBM3D  & \multirow{2}{*}{$38.78$} & \multirow{2}{*}{$34.66$} & \multirow{2}{*}{$32.38$} & \multirow{2}{*}{$30.82$} & \multirow{2}{*}{$26.13$} & $3.89$
& \multirow{2}{*}{$38.05$} & \multirow{2}{*}{$34.06$} & \multirow{2}{*}{$31.89$} & \multirow{2}{*}{$30.42$} & \multirow{2}{*}{$25.88$} & $2.11$
\\
\cite{Dabov2007} & & & & & & ($1.41$) 
& & & & & & ($1.03$) 
\\
\hline
sKSVD& \multirow{2}{*}{$41.27$} & \multirow{2}{*}{$37.37$} & \multirow{2}{*}{$35.15$} & \multirow{2}{*}{$33.59$} & \multirow{2}{*}{$28.79$} & $1.20$
& \multirow{2}{*}{$38.87$} & \multirow{2}{*}{$34.95$} & \multirow{2}{*}{$32.80$} & \multirow{2}{*}{$31.33$} & \multirow{2}{*}{$26.89$} & $1.21$
\\
\cite{elad2010sksvd} & & & & & & $(0.34)$
& & & & & & ($0.38$)
\\
\hline
\multirow{2}{*}{3D DCT}  & \multirow{2}{*}{$41.26$} & \multirow{2}{*}{$37.14$} & \multirow{2}{*}{$34.73$} & \multirow{2}{*}{$33.03$} & \multirow{2}{*}{$27.59$} & $1.69$
& \multirow{2}{*}{$38.01$} & \multirow{2}{*}{$33.60$} & \multirow{2}{*}{$30.44$} & \multirow{2}{*}{$28.50$} & \multirow{2}{*}{$22.31$} & $3.60$
\\
 & & & & & & ($0.78$) 
& & & & & & ($1.28$)
\\
\hline
VBM3D & \multirow{2}{*}{$41.10$} & \multirow{2}{*}{$37.82$} & \multirow{2}{*}{$35.78$} & \multirow{2}{*}{$34.25$} & \multirow{2}{*}{$28.65$} & $0.92$
& \multirow{2}{*}{$39.20$} & \multirow{2}{*}{$35.75$} & \multirow{2}{*}{$33.87$} & \multirow{2}{*}{$32.49$} & \multirow{2}{*}{$26.51$} & $0.61$
\\
\cite{vbm3d} & & & & & & ($0.72$) 
& & & & & & ($0.51$)
\\
\hline
VBM4D & \multirow{2}{*}{$41.42$} & \multirow{2}{*}{$37.59$} & \multirow{2}{*}{$35.30$} & \multirow{2}{*}{$33.64$} & \multirow{2}{*}{$27.76$} & $1.30$
& \multirow{2}{*}{$39.37$} & \multirow{2}{*}{$35.73$} & \multirow{2}{*}{$33.70$} & \multirow{2}{*}{$32.24$} & \multirow{2}{*}{$26.68$} & $0.63$
\\
\cite{Maggioni2012} & & & & & & ($0.86$) 
& & & & & & ($0.49$)
\\
\hline
\multirow{2}{*}{VIDOSAT} & \multirow{2}{*}{$41.94$} & \multirow{2}{*}{$38.32$} & \multirow{2}{*}{$36.13$} & \multirow{2}{*}{$34.60$} & \multirow{2}{*}{$29.87$} & $0.27$
& \multirow{2}{*}{$39.56$} & \multirow{2}{*}{$35.75$} & \multirow{2}{*}{$33.54$} & \multirow{2}{*}{$31.98$} & \multirow{2}{*}{$27.29$} & $0.55$
\\
 & & & & & & ($0.13$) 
& & & & & & ($0.29$)
\\
\hline
\textbf{VIDOSAT} & \multirow{2}{*}{$\mathbf{42.22}$} & \multirow{2}{*}{$\mathbf{38.57}$} & \multirow{2}{*}{$\mathbf{36.42}$} & \multirow{2}{*}{$\mathbf{34.88}$} & \multirow{2}{*}{$\mathbf{30.09}$} & \multirow{2}{*}{$0$}
& \multirow{2}{*}{$\mathbf{39.95}$} & \multirow{2}{*}{$\mathbf{36.11}$} & \multirow{2}{*}{$\mathbf{34.05}$} & \multirow{2}{*}{$\mathbf{32.60}$} & \multirow{2}{*}{$\mathbf{28.15}$} & \multirow{2}{*}{$0$}  
\\
\textbf{-BM} & & & & & &  
& & & & & & 
\\
\hline
\end{tabular}
\vspace{0.1in}
\caption{Comparison of video denoising PSNR values (in dB), averaged over the \textbf{ASU} dataset (left) and the \textbf{LASIP} dataset (right), for the proposed VIDOSAT, VIDOSAT-BM, and other competing methods. For each dataset and noise level, the best denoising PSNR is marked in bold. For each method, we list $\Delta$ PSNR, which denotes the average PSNR difference (with its standard deviation included in parentheses) relative to the proposed VIDOSAT-BM (highlighted in bold).}
\label{tab:denoisingDataset}
\vspace{-0.15in}
\end{table*}

\subsection{Computational Costs} \label{sec43}
In Algorithm \textbf{A1}, the computational cost of the sparse coding step is dominated by the computation of matrix-vector multiplication $\hat{\mathbf{W}} \mathbf{u}_i$, 
which scales as $O(M n^2)$ \cite{sai2015onlineTL,wen2015vidosat} for each mini-batch. 
The cost of mini-batch transform update step is $O(n^3 + Mn^2)$, which is dominated by full SVD and matrix-matrix multiplications.
The cost of the 3D denoised patch reconstruction step also scales as $O(n^3 + Mn^2)$ per mini-batch, which is dominated by the computation of matrix inverse $\hat{\mathbf{W}}^{-1}$ and multiplications.
As all overlapping patches from a $a \times b \times T$ video are sequentially processed, the computational cost of Algorithm \textbf{A1} scales as $O(abTn^3 / M + abTn^2)$.
We set $M = 15n$ in practice, so that the cost of \textbf{A1} scales as $O(abTn^2)$.
The cost of the additional BM step in Algorithm \textbf{A2} scales as $O(abTmh_1h_2)$, where $h_1 \times h_2$ is the search window size.
Therefore, the total cost of \textbf{A2} scales as $O(abTn^2 + abTmh_1h_2)$, which is on par with the state-of-the-art video denoising algorithm VBM3D \cite{vbm3d}, which is not an online method.


\begin{figure*}[t]
\begin{center}
\begin{tabular}{cccc}
\includegraphics[width=1.6in]{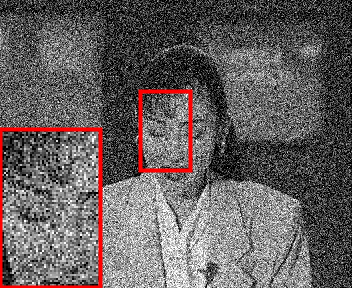} &
\includegraphics[width=1.6in]{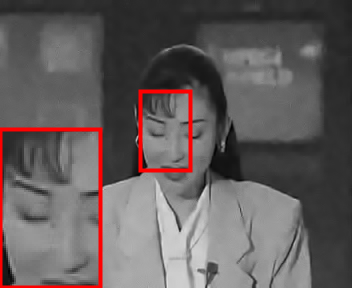} &
\includegraphics[width=1.6in]{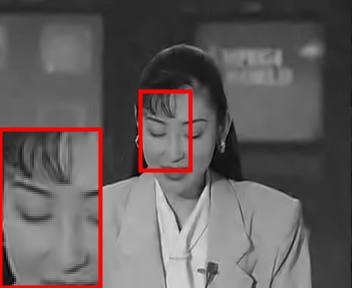} & 
\includegraphics[width=1.6in]{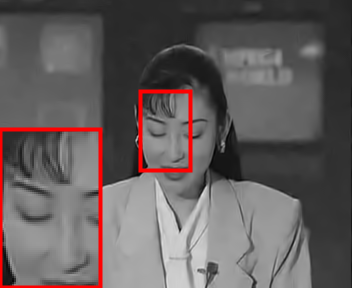} \\
{\small (a) Noisy} & {\small (c) VBM3D ($33.30$ dB)} & {\small (e) VIDOSAT ($35.84$ dB)} & {\small (g) VIDOSAT-BM ($36.11$ dB)} \\
\includegraphics[width=1.6in]{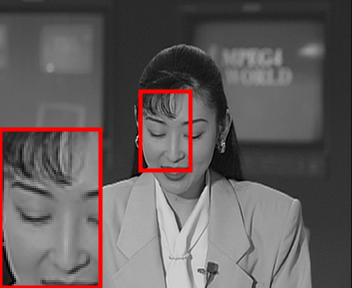} &
\includegraphics[width=1.6in]{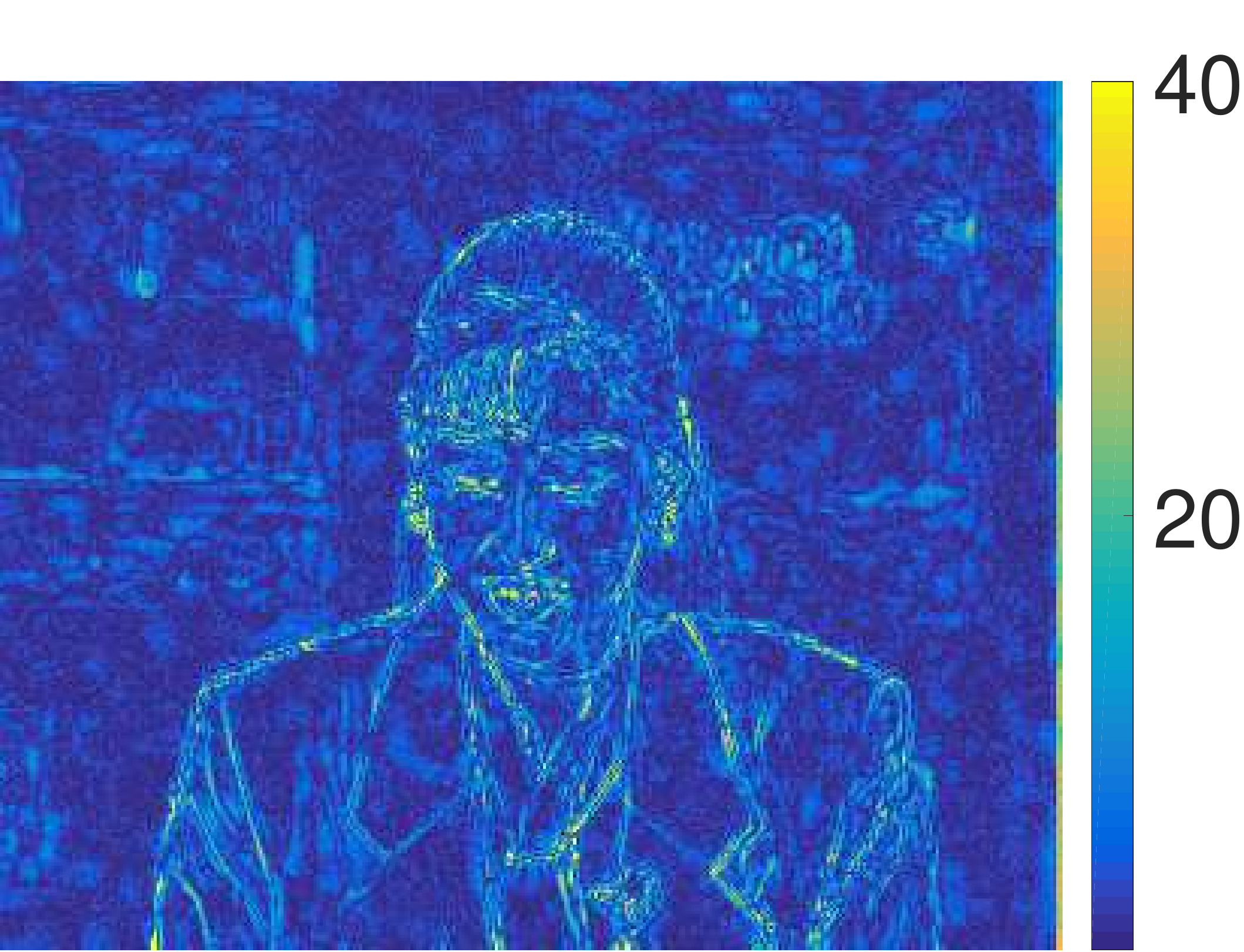} &
\includegraphics[width=1.6in]{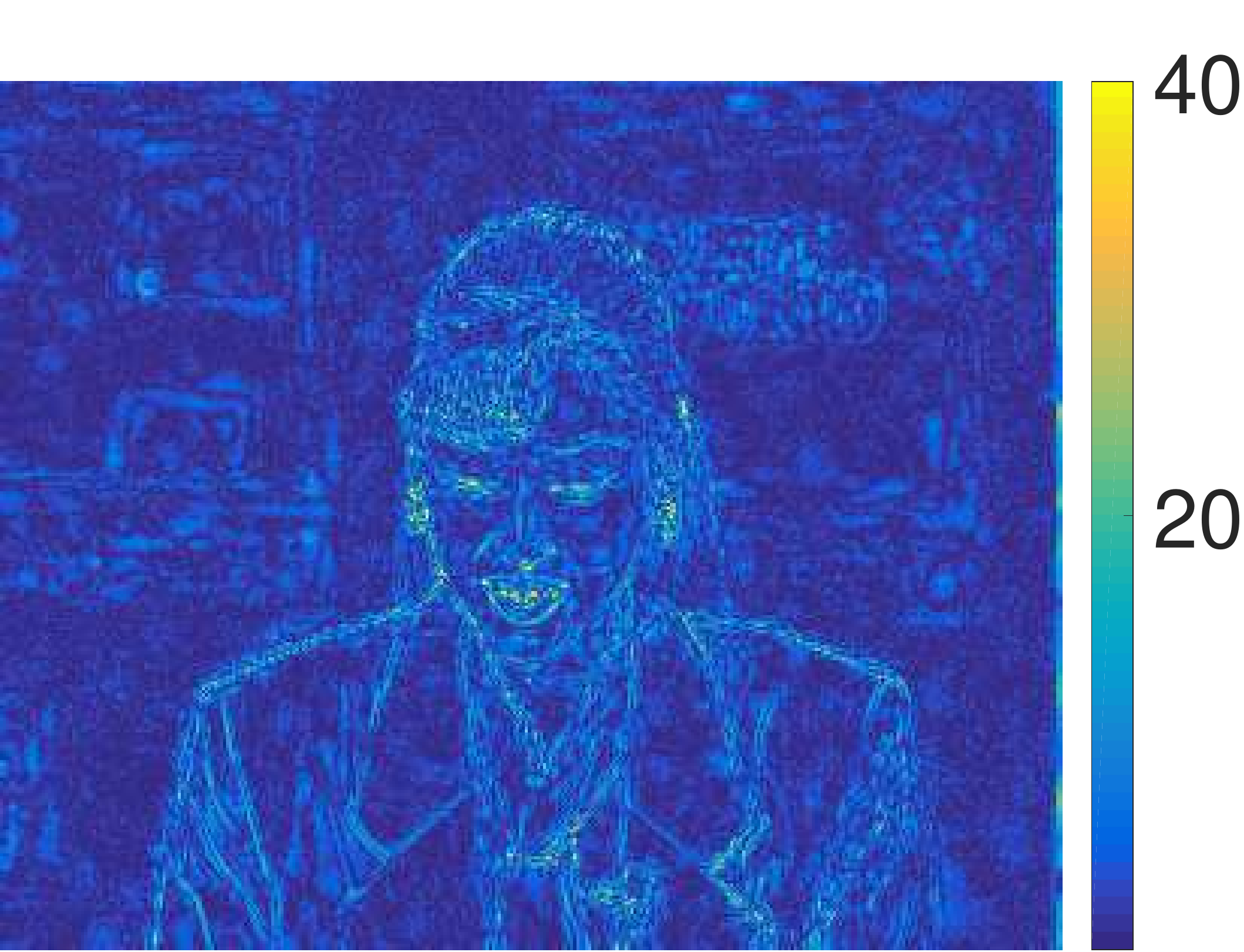} & 
\includegraphics[width=1.6in]{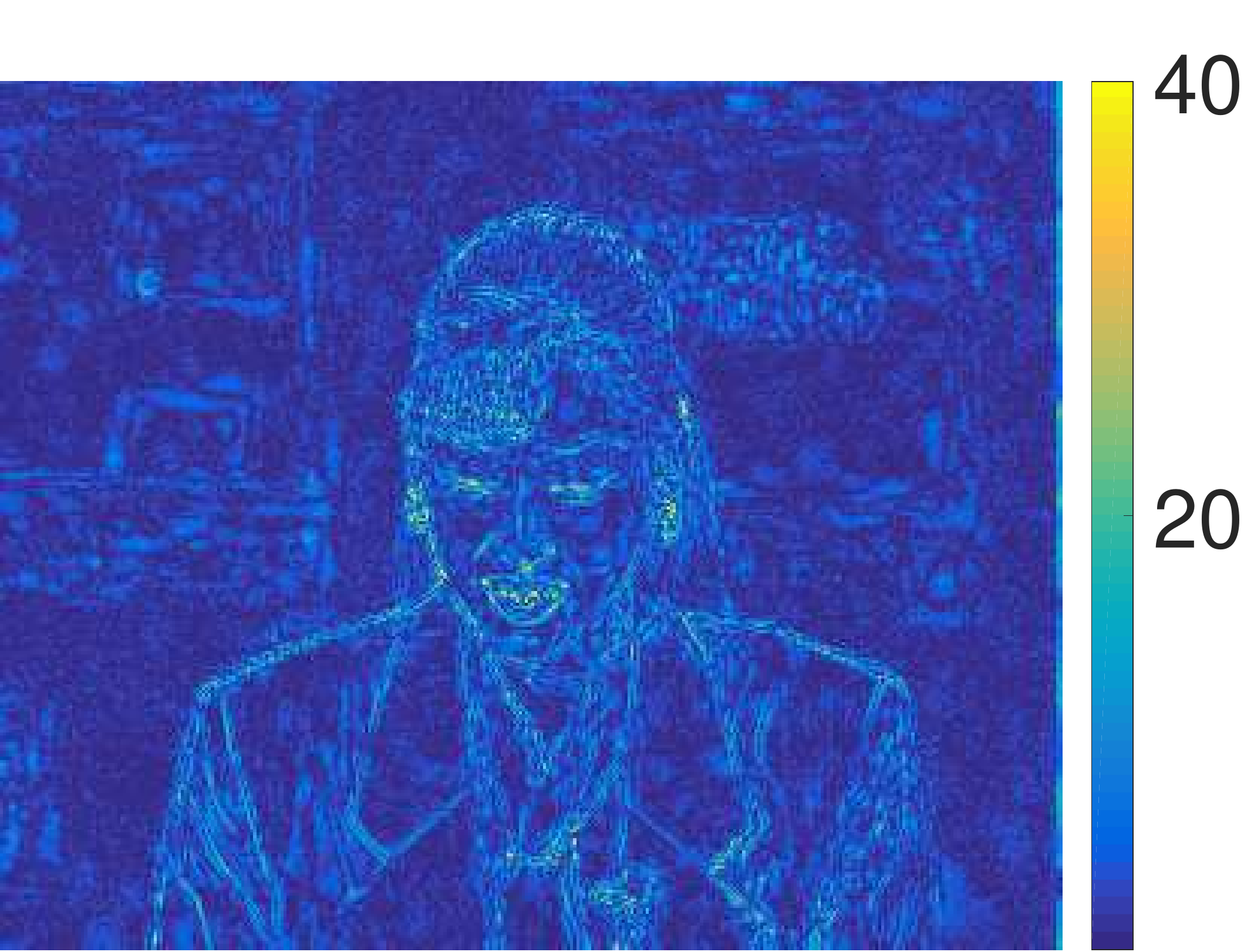} \\
{\small (b) Original} & {\small (d)} & {\small (f)} & {\small (h)}
\end{tabular}
\end{center}
\vspace{-0.1in}
\caption{(a) The noisy version ($\sigma = 50$) of (b) one frame of the \textit{Akiyo} ($288 \times 352 \times 300$) video. We show the comparison of the denoising results (resp. the magnitude of error in the denoised frame) using (c) VBM3D ($33.30$ dB), (e) VIDOSAT ($35.84$ dB) and (g) VIDOSAT-BM ($36.11$ dB) (resp. (d), (f) and (h)). The PSNR of the denoised frame is shown in the parentheses. The zoom-in region is highlighted using red box.}
\label{fig:denoisingExp1}
\end{figure*}

\begin{figure*}
\begin{center}
\begin{tabular}{cccc}
\includegraphics[width=1.6in]{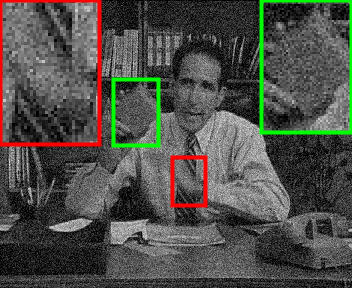} &
\includegraphics[width=1.6in]{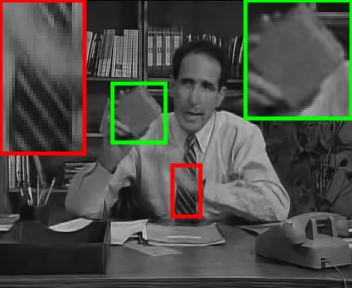} &
\includegraphics[width=1.6in]{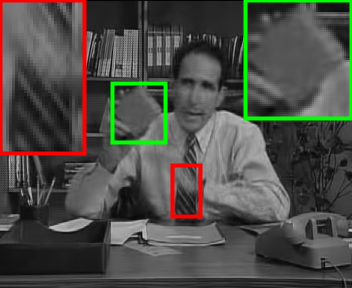} & 
\includegraphics[width=1.6in]{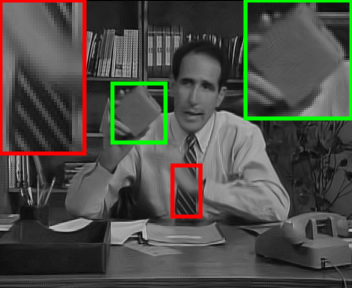} \\
{\small (a) Noisy} & {\small (c) VBM4D ($33.04$ dB)} & {\small (e) VIDOSAT ($33.43$ dB)} & {\small (g) VIDOSAT-BM ($34.01$ dB)} \\
\includegraphics[width=1.6in]{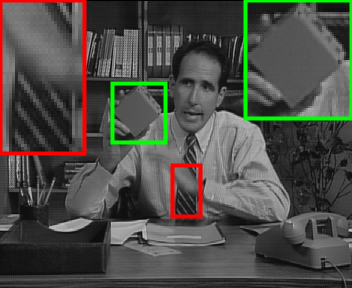} &
\includegraphics[width=1.6in]{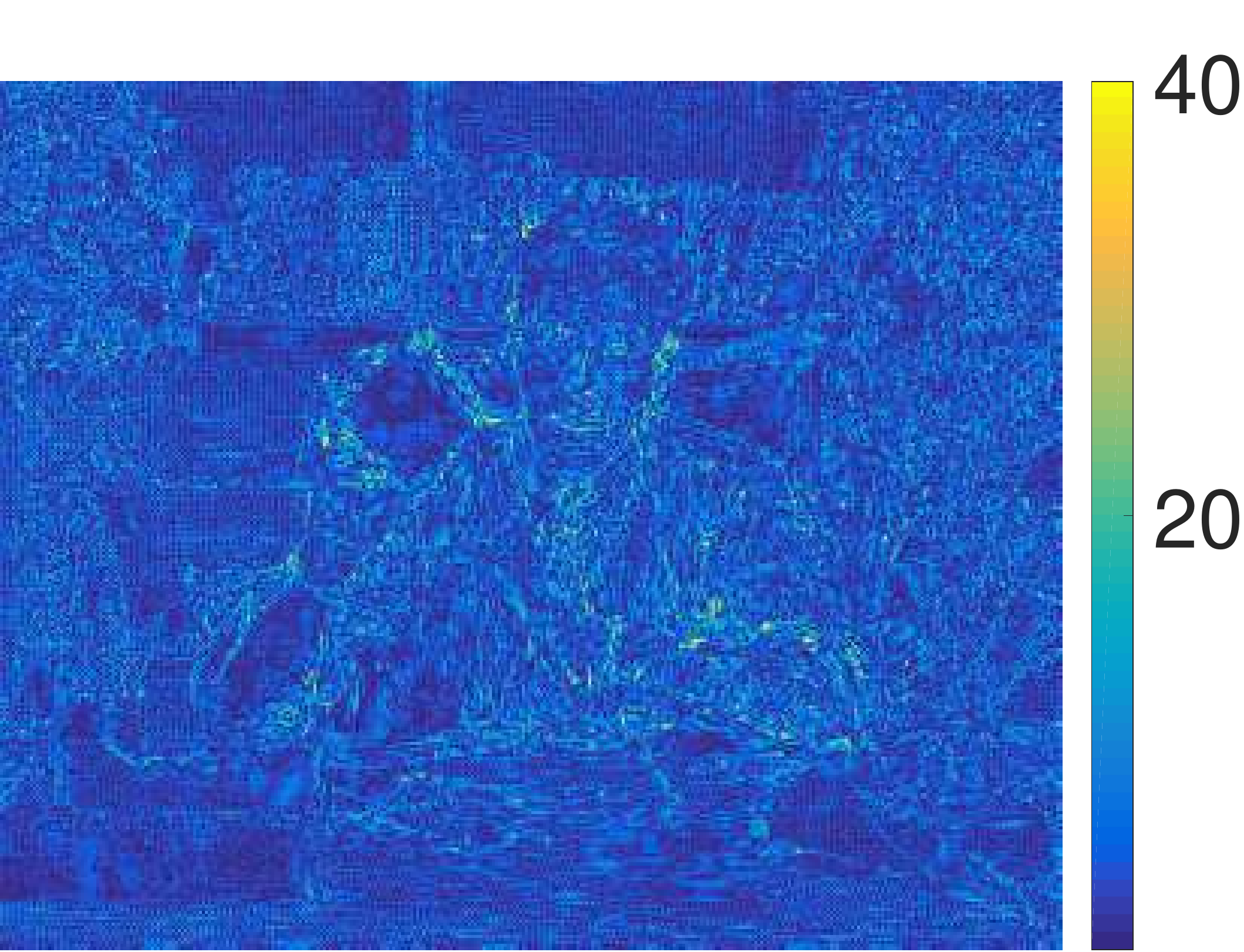} &
\includegraphics[width=1.6in]{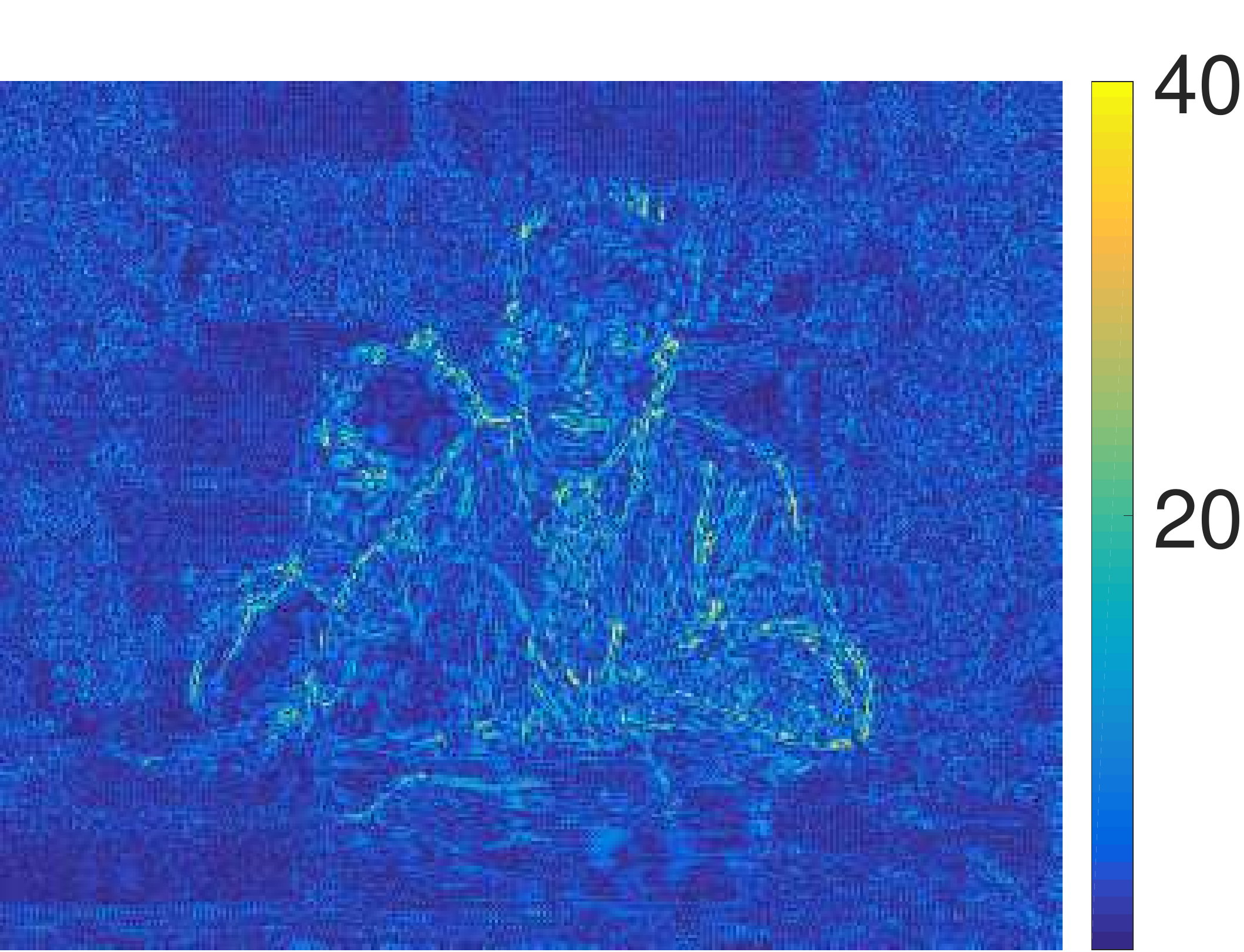} & 
\includegraphics[width=1.6in]{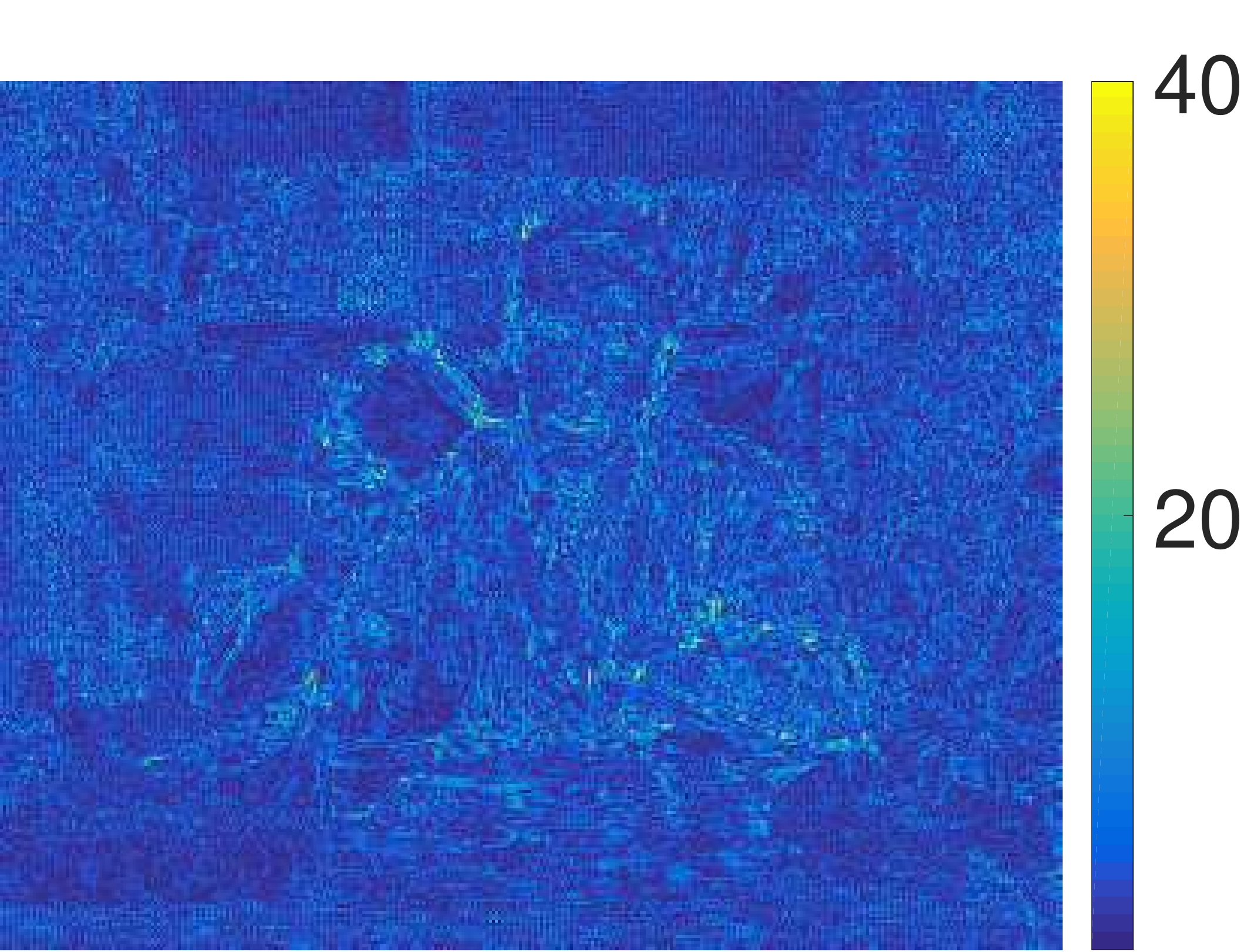} \\
{\small (b) Original} & {\small (d)} & {\small (f)} & {\small (h)}
\end{tabular}
\end{center}
\vspace{-0.1in}
\caption{(a) The noisy version ($\sigma = 20$) of (b) one frame of the \textit{Salesman} ($288 \times 352 \times 50$) video. We show the comparison of the denoising results (resp. the magnitude of error in the denoised frame) using (c) VBM4D ($33.04$ dB), (e) VIDOSAT ($33.43$ dB) and (g) VIDOSAT-BM ($34.01$ dB) (resp. (d), (f) and (h)). The PSNR of the denoised frame is shown in the parentheses. The zoom-in regions are highlighted using red and green boxes.}
\label{fig:denoisingExp2}
\vspace{-0.1in}
\end{figure*}

\begin{figure*}
\begin{center}
\begin{tabular}{cccc}
\includegraphics[width=1.6in]{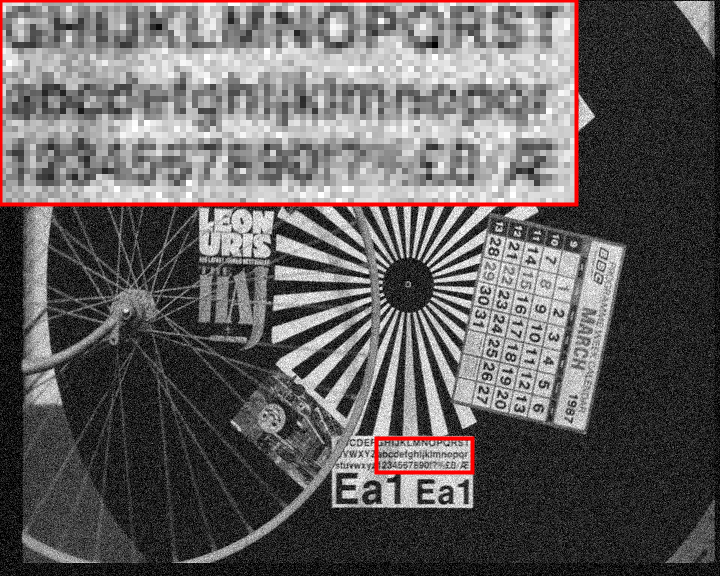} &
\includegraphics[width=1.6in]{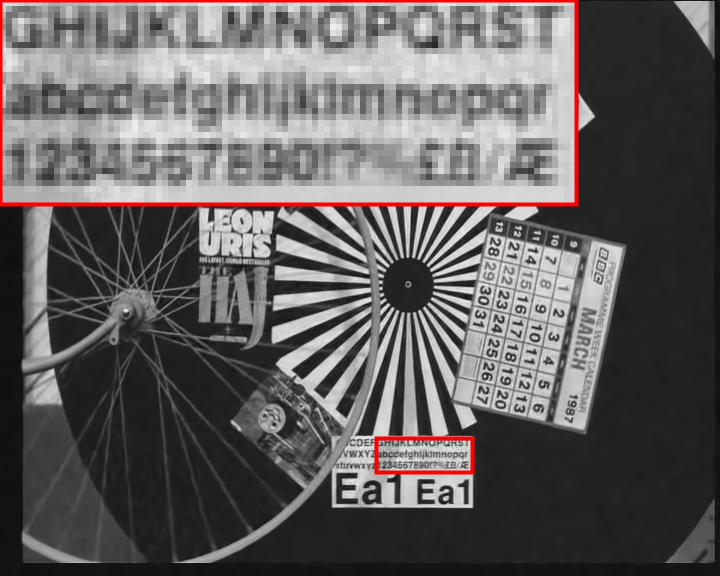} &
\includegraphics[width=1.6in]{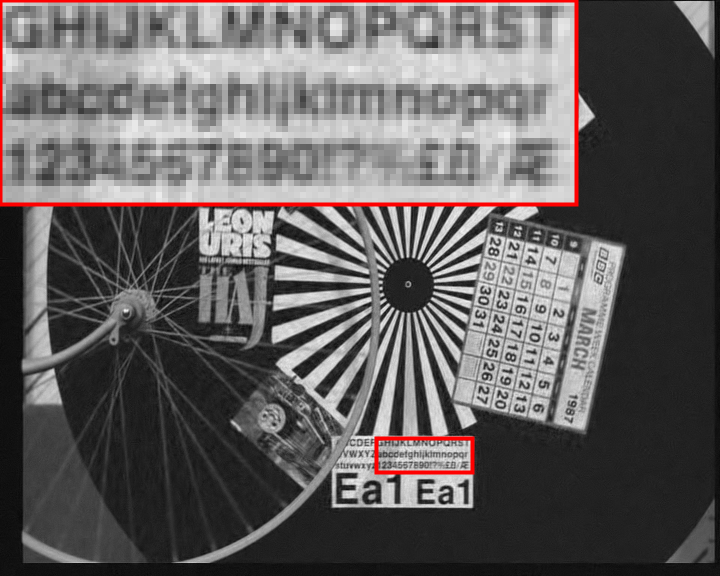} & 
\includegraphics[width=1.6in]{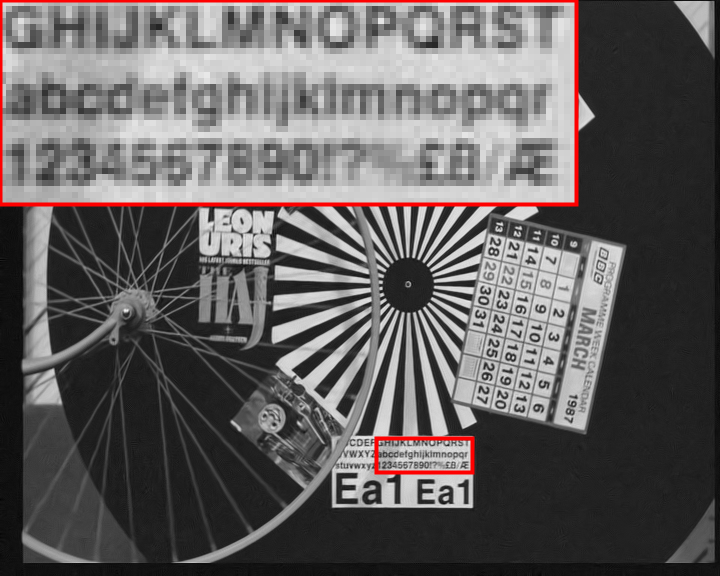} \\
{\small (a) Noisy} & {\small (c)  VBM4D ($34.00$ dB)} & {\small (e) VIDOSAT ($32.07$ dB)} & {\small (g) VIDOSAT-BM ($35.33$ dB)} \\
\includegraphics[width=1.6in]{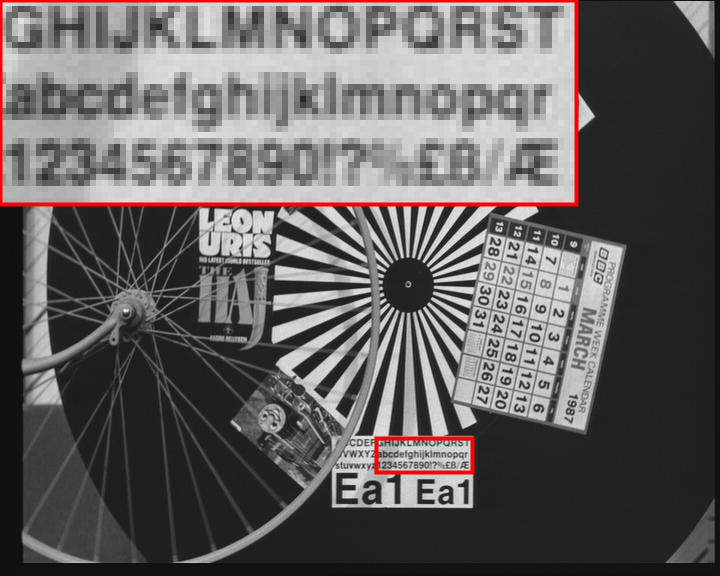} &
\includegraphics[width=1.6in]{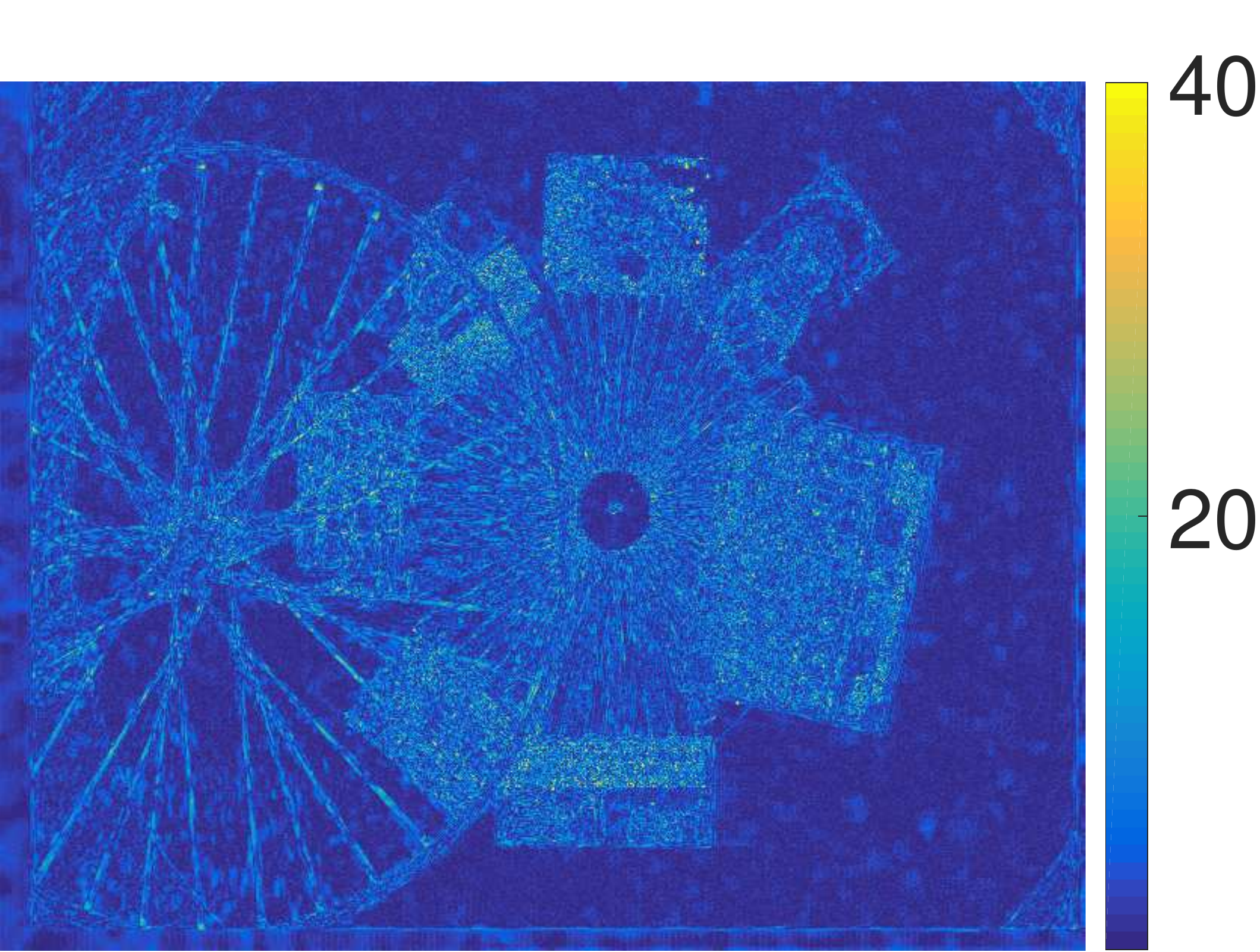} &
\includegraphics[width=1.6in]{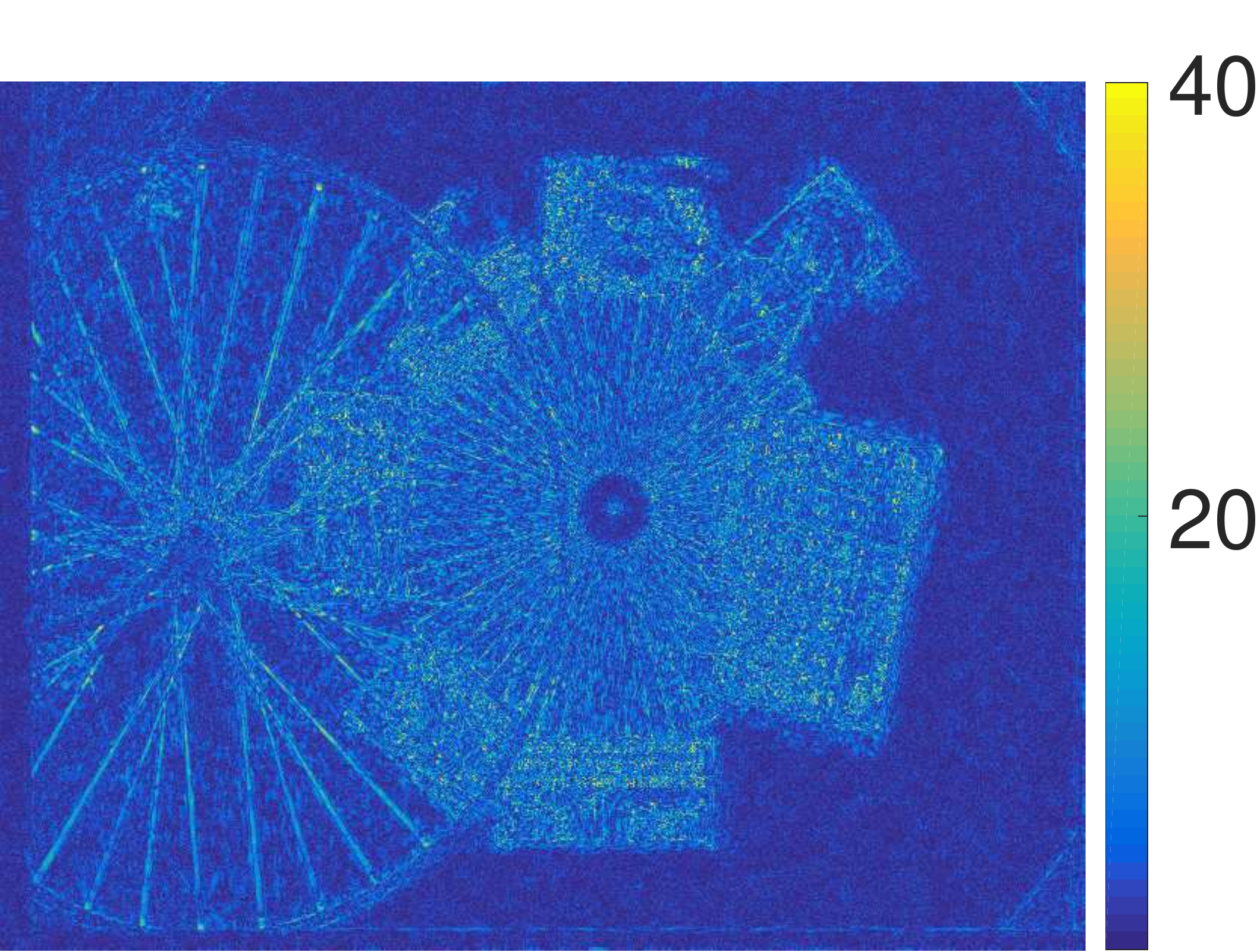} & 
\includegraphics[width=1.6in]{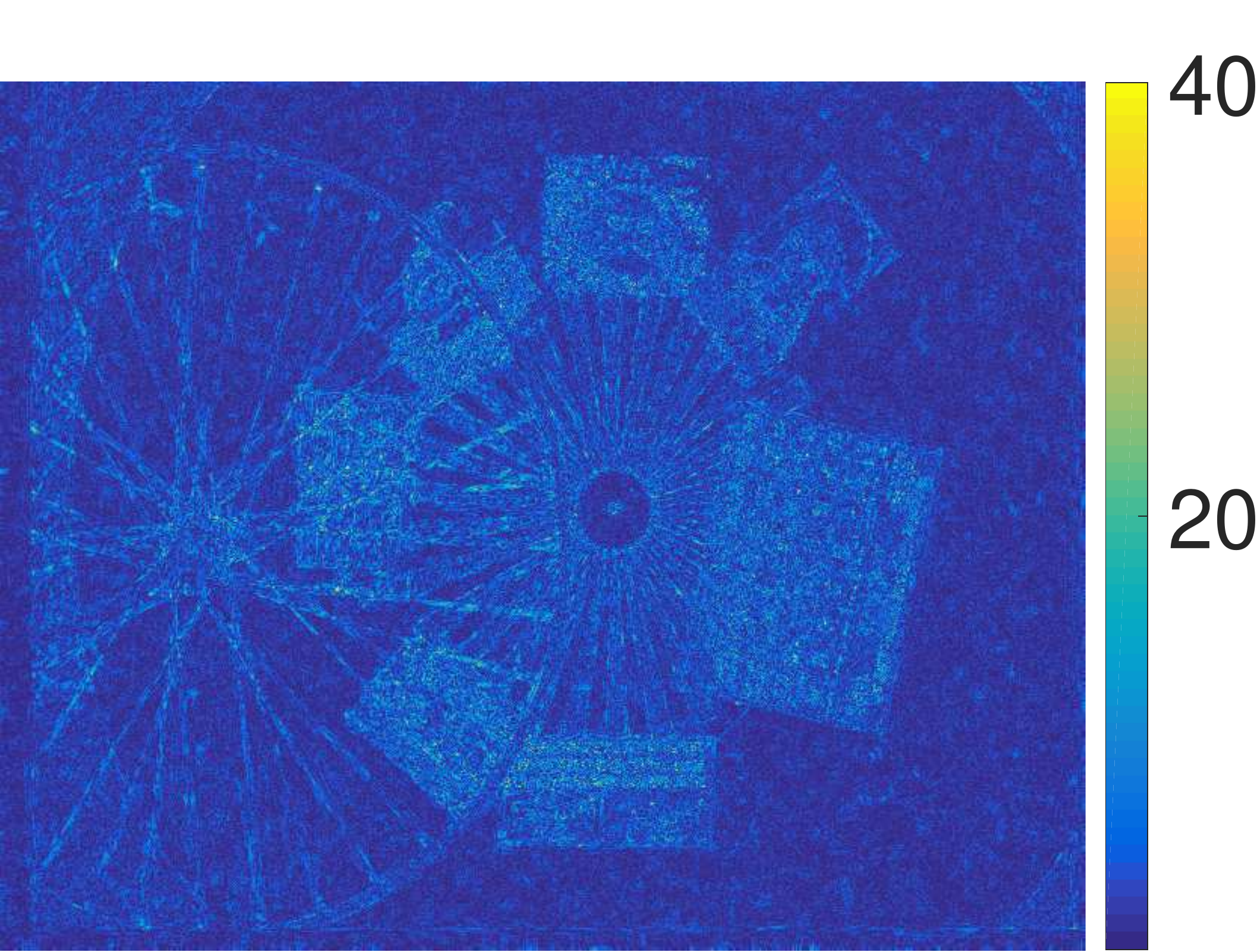} \\
{\small (b) Original} & {\small (d)} & {\small (f)} & {\small (h)}
\end{tabular}
\end{center}
\vspace{-0.1in}
\caption{(a) The noisy version ($\sigma = 20$) of (b) one frame of the \textit{Bicycle} ($576 \times 720 \times 30$) video. We show the comparison of the denoising results (resp. the magnitude of error in the denoised frame) using (c) VBM4D ($34.00$ dB), (e) VIDOSAT ($32.07$ dB) and (g) VIDOSAT-BM ($35.33$ dB) (resp. (d), (f) and (h)). The PSNR of the denoised frame is shown in the parentheses. The zoom-in region is highlighted using red box.}
\label{fig:denoisingExp3}
\end{figure*}

\begin{figure*}
\begin{center}
\begin{tabular}{ccc}
\includegraphics[width=2.2in]{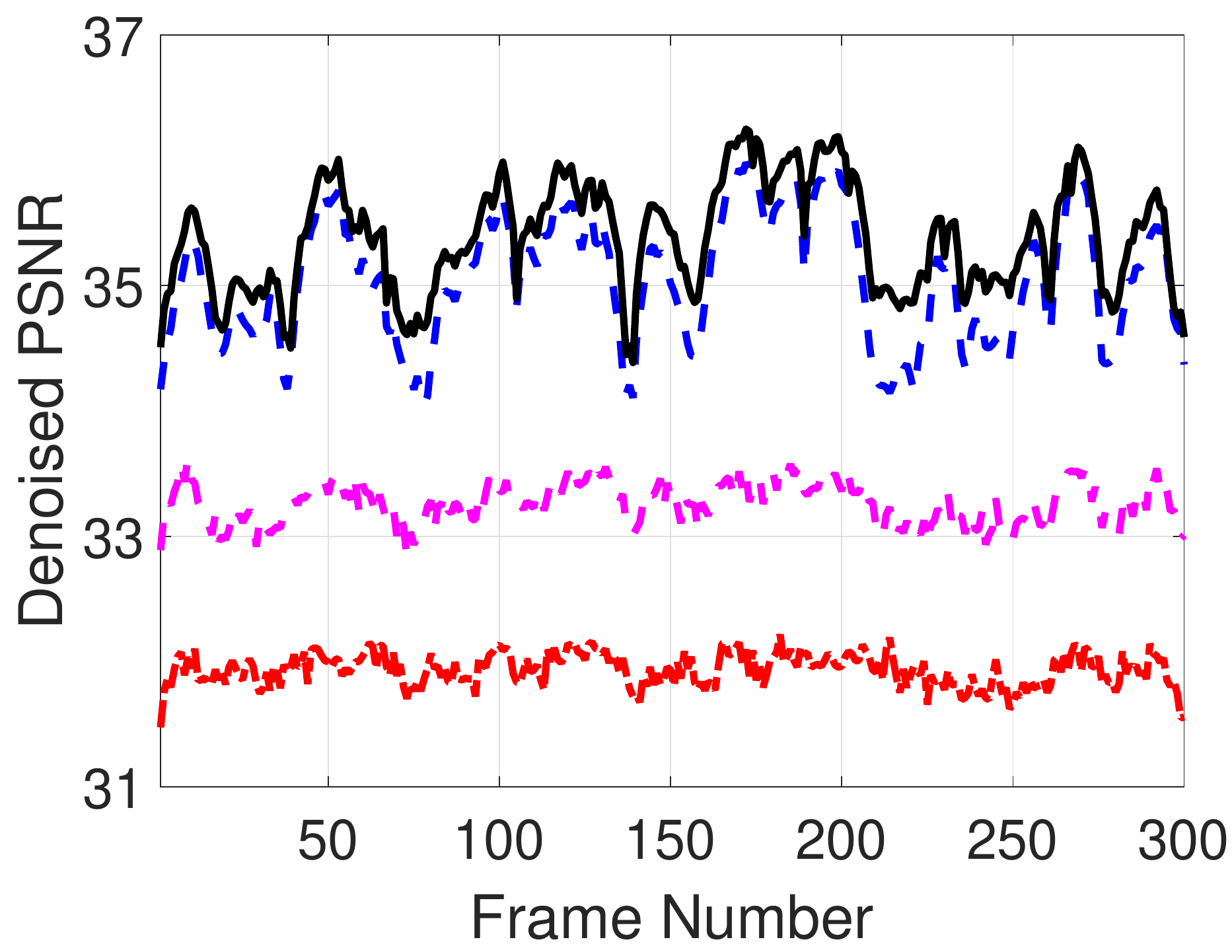} &
\includegraphics[width=2.2in]{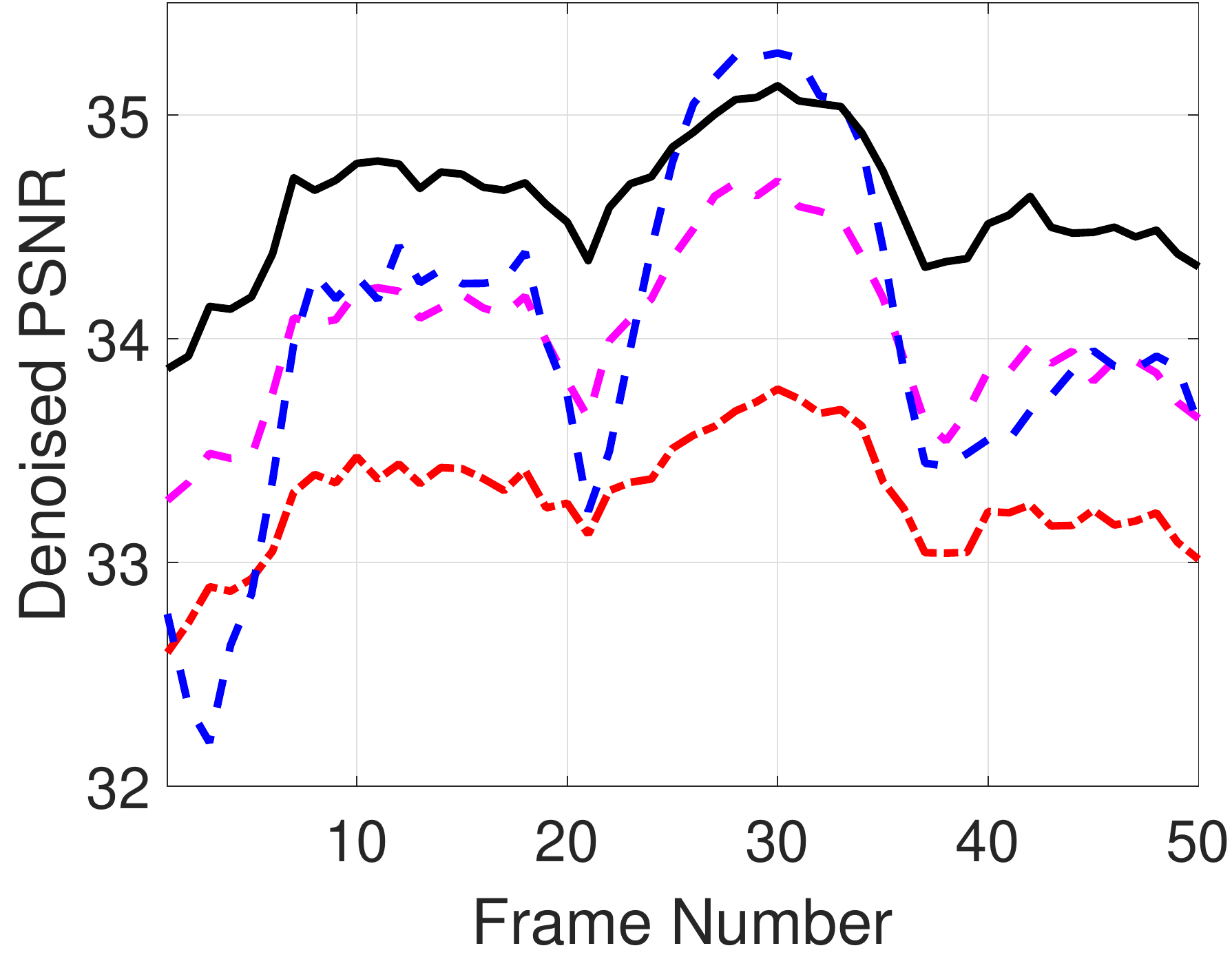} &
\includegraphics[width=2.2in]{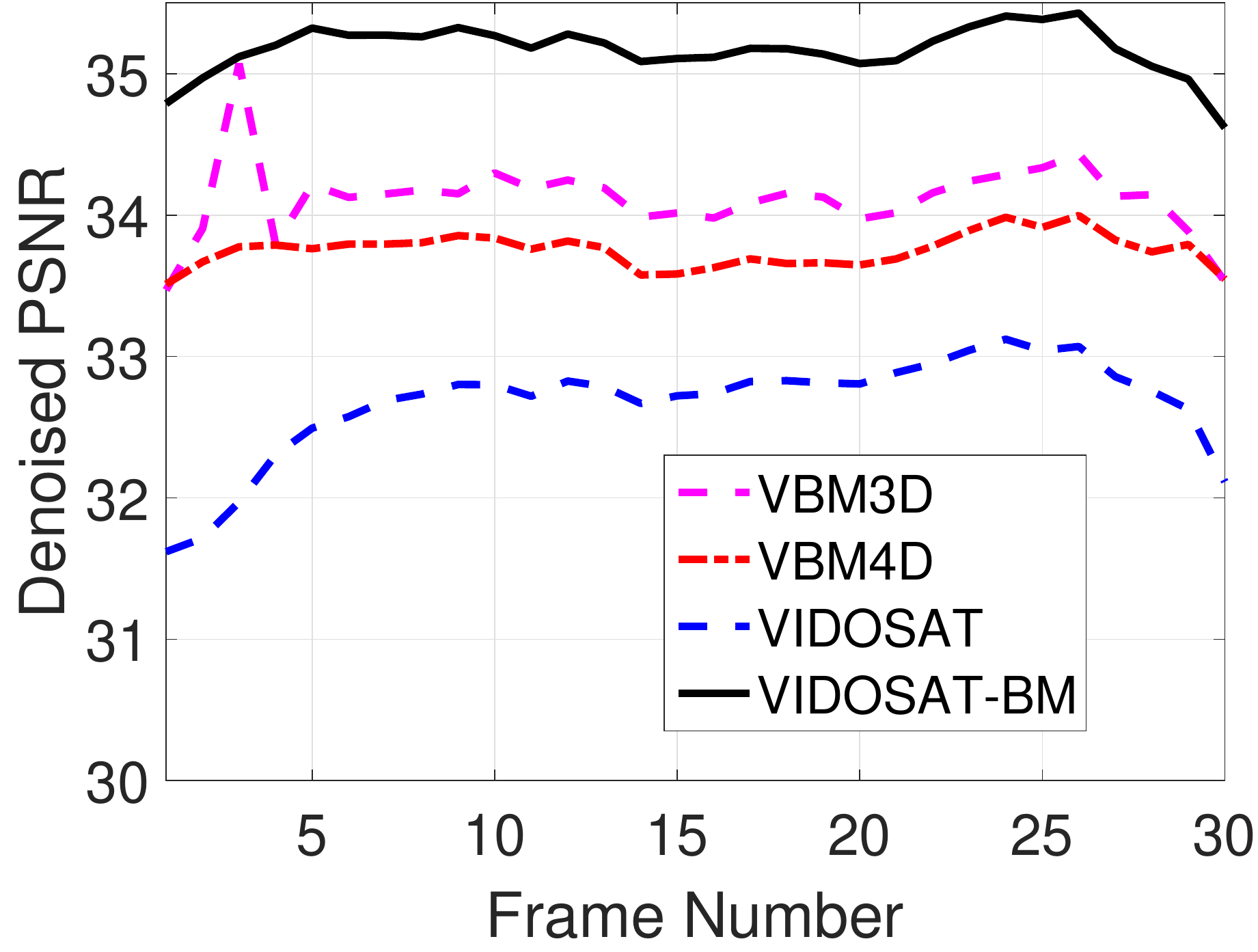} \\
{\small (a)} & {\small (b)} & {\small (c)}
\end{tabular}
\end{center}
\vspace{-0.1in}
\caption{Frame-by-frame PSNR (dB) for (a) \textit{Akiyo} with $\sigma = 50$, (b) \textit{Salesman} with $\sigma = 20$, and (c) \textit{Bicycle} with $\sigma = 20$, denoised by VBM3D, VBM4D, and the proposed VIDOSAT and VIDOSAT-BM schemes, respectively.}
\label{fig:framePSNR}
\vspace{-0.1in}
\end{figure*}

\section{Experiments} \label{sec5}

\subsection{Implementation and Parameters} \label{sec51}

\subsubsection{Testing Data} \label{sec511}
We present experimental results demonstrating the promise of the proposed VIDOSAT and VIDOSAT-BM online video denoising methods. We evaluate the proposed algorithms by denoising all $34$ videos from $2$ public datasets, including $8$ videos from the LASIP video dataset \footnote{Available at \url{http://www.cs.tut.fi/~lasip/foi_wwwstorage/test_videos.zip}} \cite{vbm3d,Maggioni2012}, and $26$ videos the Arizona State University (ASU) Video Trace Library \footnote{Available at \url{http://trace.eas.asu.edu/yuv/}. Only videos with less than $1000$ frames are selected for our image denoising experiments.} \cite{seeling2014video}.
The testing videos contain $50$ to $870$ frames, with the frame resolution ranging from $176 \times 144$ to $720 \times 576$. 
Each video involves different types of motion, including translation, rotation, scaling (zooming), etc. 
The color videos are all converted to gray-scale. 
We simulated i.i.d. zero-mean Gaussian noise at 5 different noise levels (with standard deviation $\sigma = 5$, $10$, $15$, $20$, and $50$) for each video.

\subsubsection{Implementation Details} \label{sec512}
We include several minor modifications of VIDOSAT and VIDOSAT-BM algorithms for improved performance.
At each time instant $t$, we perform multiple passes of denoising for each $ {\mathcal{Y}}_t$, by iterating over Steps $(1)$ to $(5)$ multiple times. 
In each pass, we denoise the output from the previous iteration \cite{wen2015octobos,sai2015onlineTL}. 
As the sparsity penalty weights are set proportional to the noise level, $\alpha_{j, i} = \alpha_{0} \sigma$, the noise standard deviation $\sigma$ in each such pass is set to an empirical estimate \cite{wen2015vidosat,wen2015octobos} of the remaining noise in the denoised frames from the previous pass.
These multiple passes, although increasing the computation in the algorithm, do not increase the inherent latency $m-1$ of the single pass algorithm described earlier. 

The following details are specifically for VIDOSAT-BM. 
First, instead of performing BM over the noisy input buffer $ {\mathcal{Y}}_t$, we pre-clean $ {\mathcal{Y}}_t$ using the VIDOSAT mini-batch denoising Algorithm \textbf{A1}, and then perform BM over the VIDOSAT denoised output. 
Second, when denoised 3D patches are aggregated to the output buffer, we assign them different weights, which are proportional to the sparsity level of their optimal sparse codes \cite{wen2017strollr}. 
The weights are also accumulated and used for the output normalization. 

\subsubsection{Hyperparameters} \label{sec513}
We work with fully overlapping patches (spatial patch stride of $1$ pixel) with spatial size $n_1 = n_2 = 8$, and temporal depth of $m = 9$ frames, which also corresponds to the depth of buffer $\mathcal{Y}$. 
It follows that for a video with $N_1 \times N_2$ frames, the buffer $\mathcal{Y}$ contains $m N_1 N_2$ pixels, and $P= (N_1-n_1+1)(N_2-n_2+1)$ 3D patches.
We set the sparsity penalty weight parameter $\alpha_0 = 1.9$, the transform regularizer weight constant $\lambda_0 = 10^{-2}$, and the mini-batch size $M = 15 \times mn_1n_2$. The transform $\mathbf{W}$ is initialized with the 3D DCT $\mathbf{W}_0$. 
For the other parameters, we adopt the settings in prior works \cite{sai2015onlineTL,wen2015vidosat,wen2015octobos}, such as the forgetting factor $\rho = 0.68, 0.72, 0.76, 0.83, 0.89$, and the number of passes $L_p = 1, 2, 3, 3, 4$ for $\sigma = 5, 10, 15, 20, 50$, respectively. 
The values of $\rho$ and $L_p$ both increase as the noise level increases.
The larger $\rho$ helps prevent overfitting to noise, and the larger number of pass improves denoising performance at higher noise level.
For VIDOSAT-BM, we set the local search window size $h_1 = h_2 = 21$.

\subsection{Video Denoising Results} \label{sec52}

\subsubsection{Competing Methods} \label{sec521}
We compare the video denoising results obtained using the proposed VIDOSAT and VIDOSAT-BM algorithms to several well-known alternatives, 
including the frame-wise BM3D denoising method (fBM3D) \cite{Dabov2007}, the image sequence denoising method using sparse KSVD (sKSVD) \cite{elad2010sksvd}, VBM3D \cite{vbm3d} and VBM4D methods \cite{Maggioni2012}. 
We used the publicly available implementations of these methods. 
Among these competing methods, fBM3D denoises each frame independently by applying a popular BM3D image denoising method; 
sKSVD exploits adaptive spatio-temporal sparsity but the dictionary is not learned online; 
and VBM3D and VBM4D are popular and state-of-the-art video denoising methods. 
Moreover, to better understand the advantages of the online high-dimensional transform learning, we apply the proposed video denoising framework, but fixing the sparsifying transform in VIDOSAT to 3D DCT, which is referred as the 3D DCT method.

\subsubsection{Denoising Results} \label{sec522}

We present video denoising results using the proposed VIDOSAT and VIDOSAT-BM algorithms, 
as well as using the other aforementioned competing methods.
To evaluate the performance of the various denoising schemes, we measure the peak signal-to-noise ratio (PSNR) in decibels (dB), which is computed between
the noiseless reference and the denoised video.

Table \ref{tab:denoisingDataset} lists the video denoising PSNRs obtained by the two proposed VIDOSAT methods, as well as the five competing methods. 
It is clear that the proposed VIDOSAT and VIDOSAT-BM approaches both generate better denoising results with higher average PSNR values, 
compared to the competing methods.
The VIDOSAT-BM denoising method provides average PSNR improvements (averaged over all $34$ testing videos from both datasets and all noise levels) of $0.9$ dB, $1.1$ dB, $1.2$ dB, $2.1$ dB, and $3.5$ dB, over the VBM3D, VBM4D, sKSVD, 3D DCT, and fBM3D denoising methods.
Importantly, VIDOSAT-BM consistently outperforms all the competing methods for all testing videos and noise levels.
Among the two proposed VIDOSAT algorithms, the average video denoising PSNR by VIDOSAT-BM is $0.3$ dB higher than that using the VIDOSAT method, thanks to the use of the block matching for modeling dynamics and motion in video.

We illustrate the denoising results and improvements provided by VIDOSAT and VIDOSAT-BM with some examples.

\begin{figure}[!t]
\centering
\includegraphics[width=3.3in]{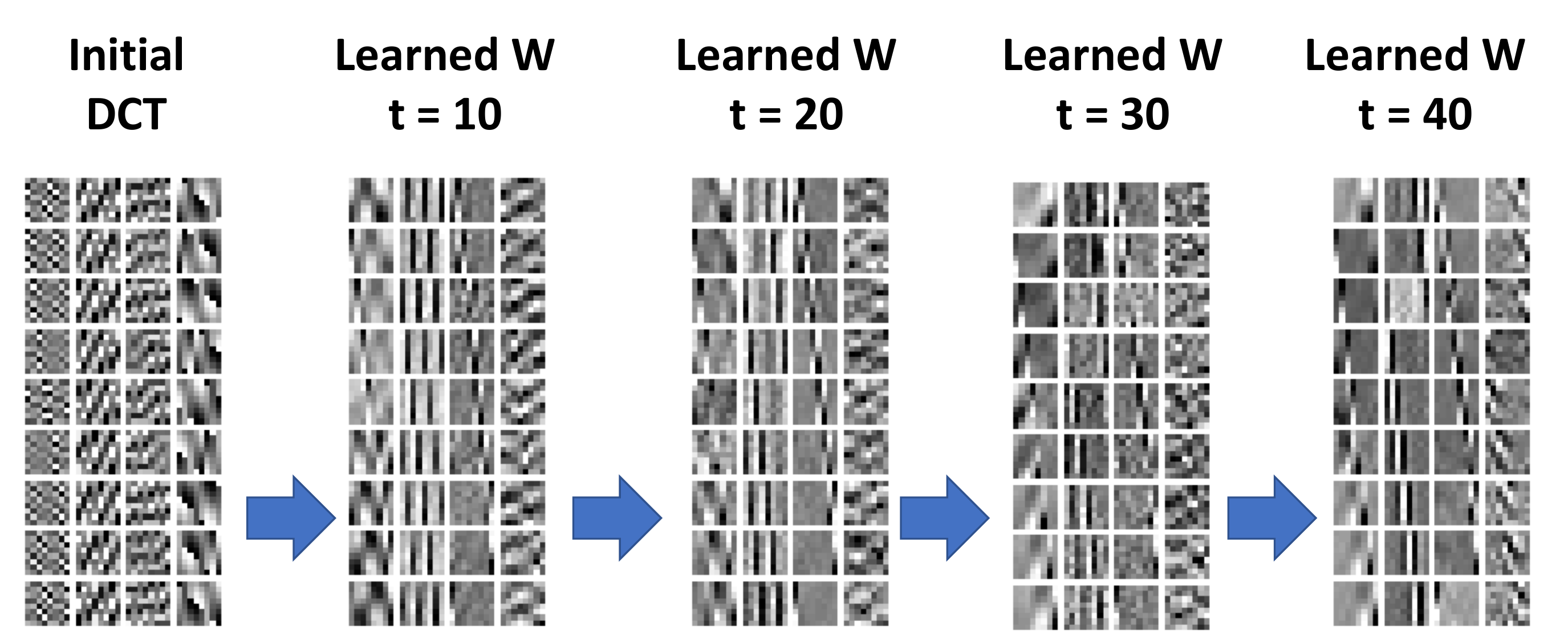} \\
(a) $\hat{\mathbf{W}}_t$ learned using VIDOSAT \\
\includegraphics[width=3.3in]{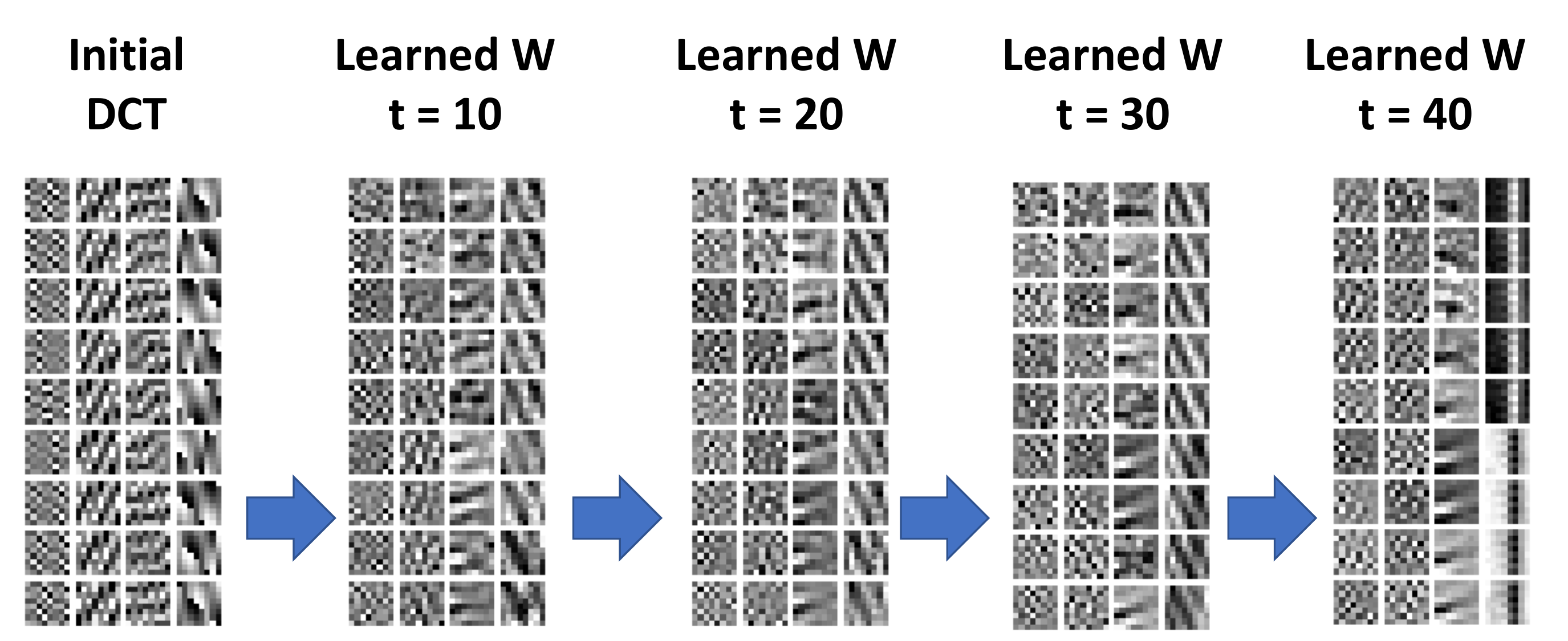} \\
(b) $\hat{\mathbf{W}}_t$ learned using VIDOSAT-BM
\vspace{-0.08in}
\caption{Example atoms (i.e., $4$ rows) of the initial 3D DCT (with depth $m = 9$), and the online learned 3D sparsifying transform using (a) VIDOSAT, and (b) VIDOSAT-BM, at times $10$ to $40$: the atoms (i.e., rows) of the learned $\hat{\mathbf{W}}$ are shown as $m = 9$ patches in each column. These $9$ patches together form the $8 \times 8 \times 9$ 3D atoms.}
\label{fig:visualW}
\vspace{-0.2in}
\end{figure}

\paragraph{Fig. \ref{fig:denoisingExp1}} shows one denoised frame of the video \textit{Akiyo} ($\sigma = 50$), which involves static background and a relatively small moving region (The magnitudes of error in Fig. \ref{fig:denoisingExp1} are clipped for viewing). 
The denoising results by VIDOSAT and VIDOSAT-BM both demonstrate similar visual quality improvements over the result by VBM3D. 
Fig. \ref{fig:framePSNR}(a) shows the frame-by-frame PSNRs of the denoised \textit{Akiyo}, in which VIDOSAT and VIDOSAT-BM provide comparable denoising PSNRs, and both outperform the VBM3D and VBM4D schemes consistently by a sizable margin.

\paragraph{Fig. \ref{fig:denoisingExp2}} shows one denoised frame of the video \textit{Salesman} ($\sigma = 20$) that involves occasional but fast movements (e.g., hand waving) in the foreground. 
The denoising result by VIDOSAT improves over the VBM4D result in general, but also shows some artifacts in regions with strong motion.
Instead, the result by VIDOSAT-BM provides the best visual quality in both the static and the moving parts.
Fig. \ref{fig:framePSNR}(b) shows the frame-by-frame PSNRs of the denoised \textit{Salesman}. 
VIDOSAT-BM provides large improvements over the other methods including VIDOSAT for most frames, and the PSNR is more stable (smaller deviations) over frames.
Fig. \ref{fig:visualW} shows example atoms (i.e., rows) of the initial 3D DCT transform, and the online learned transforms using (a) VIDOSAT and (b) VIDOSAT-BM at different times $t$.
For the learned $\hat{\mathbf{W}}_t$'s using both VIDOSAT and VIDOSAT-BM, their atoms are observed to gradually evolve, in order to adapt to the dynamic video content.
The learned transform atoms using VIDOSAT in Fig. \ref{fig:visualW}(a) demonstrate linear shifting structure along the patch depth $m$, which is likely to compensate the video motion (e.g., translation).
On the other hand, since the 3D patches are formed using BM in VIDOSAT-BM, such structure is not observed in Fig. \ref{fig:visualW}(b) when $\hat{\mathbf{W}}_t$ is learned using VIDOSAT-BM.

\paragraph{Fig. \ref{fig:denoisingExp3}} shows one denoised frame of the video \textit{Bicycle} ($\sigma = 20$), which contains a large area of complex movements (e.g., rotations) throughout the video. 
In this case, the denoised frame using the VIDOSAT is worse than VBM4D. 
However, VIDOSAT-BM provides superior quality compared to all the methods. 
This example demonstrates the effectiveness of joint block matching and learning in the proposed VIDOSAT-BM scheme, especially when processing highly dynamic videos.
Fig. \ref{fig:framePSNR}(c) shows the frame-by-frame PSNRs of the denoised \textit{Bicycle}, in which VIDOSAT-BM significantly improves over VIDOSAT, and also outperforms both VBM3D and VBM4D for all frames.


\section{Conclusions} \label{sec6}

We presented a novel framework for online video denoising based on efficient high-dimensional sparsifying transform learning. The transforms are learned in an online manner from spatio-temporal patches. These patches are constructed either from corresponding 2D patches of consecutive frames or using an online block matching technique.
The learned models effectively capture the dynamic changes in videos. 
We demonstrated the promising performance of the proposed video denoising schemes for several standard datasets.
Our methods outperformed all compared methods, 
which included a version of our the proposed video denoising scheme in which the learning of the sparsifying transform was eliminated and instead it was fixed to 3D DCT,
as well as denoising using learned synthesis dictionaries, and the state-of-the-art VBM3D and VBM4D methods.
While this work provides an initial study of the promise of the proposed data-driven online video denoising methodologies, we plan to study the potential implementation and acceleration of the proposed schemes for real-time video processing in future work.

\ifCLASSOPTIONcaptionsoff
  \newpage
\fi

\bibliographystyle{./IEEEtran}
\bibliography{./VplusJournal}

\end{document}